\documentclass[11pt]{article}

\usepackage[final]{acl}

\usepackage{times}
\usepackage{latexsym}
\usepackage{multicol}
\usepackage{multirow}
\usepackage{booktabs}
\usepackage{amssymb}
\usepackage{amsmath}
\usepackage{hyperref}
\usepackage{inconsolata}
\usepackage{graphicx}
\usepackage[utf8]{inputenc}
\usepackage{subcaption} 
\usepackage[most]{tcolorbox}
\newtcolorbox{promptbox}[1]{
    colback=gray!5!white,
    colframe=gray!75!black,
    fonttitle=\bfseries,
    title=#1,
    arc=2mm,
    boxrule=0.5pt,
    left=10pt,
    right=10pt,
    top=5pt,
    bottom=5pt
}

\usepackage[T1]{fontenc}

\usepackage{array}
\usepackage{fontspec}

\usepackage{polyglossia}

\setdefaultlanguage{english}
\setotherlanguages{hindi,bengali,tamil,marathi,gujarati}

\newfontfamily\hindifont[Script=Devanagari, Path=./]{NotoSerifDevanagari.ttf}
\newfontfamily\marathifont[Script=Devanagari, Path=./]{NotoSerifDevanagari.ttf}
\newfontfamily\bengalifont[Path=./, BoldFont={Kalpurush.ttf}, BoldItalicFont={Kalpurush.ttf}]{Kalpurush.ttf}
\newfontfamily\gujaratifont[ Path=./]{NotoSerifGujarati.ttf}
\newfontfamily\tamilfont[ Path=./]{NotoSerifTamil.ttf}

\usepackage[utf8]{inputenc}

\usepackage{microtype}

\title{FIND: Toward Multimodal Financial Reasoning and Question Answering for Indic Languages}

\author{Sarmistha Das$^1$$^*$, Vaibhav Vishal$^1$$^*$, Syed Ibrahim Ahmad$^1$$^*$\\
 \textbf{Manish Gupta$^2$}, \textbf{Sriparna Saha$^1$}\\
  $^1$Indian Institute of Technology Patna, India\quad $^2$Microsoft, India\\
\texttt{sarmistha1515@gmail.com, vvaibhav728@gmail.com, syediahmad0@gmail.com}\\
\texttt{gmanish@microsoft.com, sriparna@iitp.ac.in}
  }

\begin{document}
\maketitle
\begin{abstract}
Financial decision-making in multilingual settings demands accurate numerical reasoning grounded in diverse modalities, yet existing benchmarks largely overlook this high-stakes, real-world challenge, especially for Indic languages. We introduce \textit{FinVQA}, a benchmark for evaluating financial numerical and multimodal reasoning in multilingual Indic contexts. \textit{FinVQA} spans English, Hindi, Bengali, Marathi, Gujarati, and Tamil, and comprises 18,900 samples across 14 financial domains. The dataset captures diverse reasoning paradigms under realistic constraints, and is structured across three difficulty levels (easy, moderate, hard) and four question formats: multiple choice, fill-in-the-blank, table matching, and true/false. To address these challenges, we propose \textit{FIND}, a framework that combines supervised fine-tuning with constraint-aware decoding to promote faithful numerical reasoning, robust multimodal grounding, and structured decision-making. Together, \textit{FinVQA} and \textit{FIND} establish a rigorous evaluation and modeling paradigm for high-stakes multilingual multimodal financial reasoning\footnote{$^*$ These authors contributed equally.}.
\end{abstract}

\section{Introduction}
Recently, driven by advances in train-time and test-time scaling~\cite{kaplan2020scaling,openai2024learning}, large language models (LLMs) have demonstrated substantial improvements in long-horizon reasoning through structured inference and strategic deliberation \cite{yue2024mmmu,das2024wealth}. These reasoning enhanced models~\cite{ghosh2025survey}, commonly referred to as Large Reasoning Models (LRMs)~\citep{2024d,c,2025,DeepSeek-AI,2025guo,team2024,team2025kimi,Gemini2025}, have achieved strong performance on complex multi-step tasks in domains such as, advisory~\cite{das2025unlocking,das2025fin} programming~\citep{chen2021evaluating,jain2024livecodebench}, mathematics~\cite{lightman2023let,mao2024champ} and science~\cite{lu2023mathvista,wang2023scibench,yue2024mmmu}. 

\begin{figure*}[h]
    \centering
    \includegraphics[width=\linewidth]{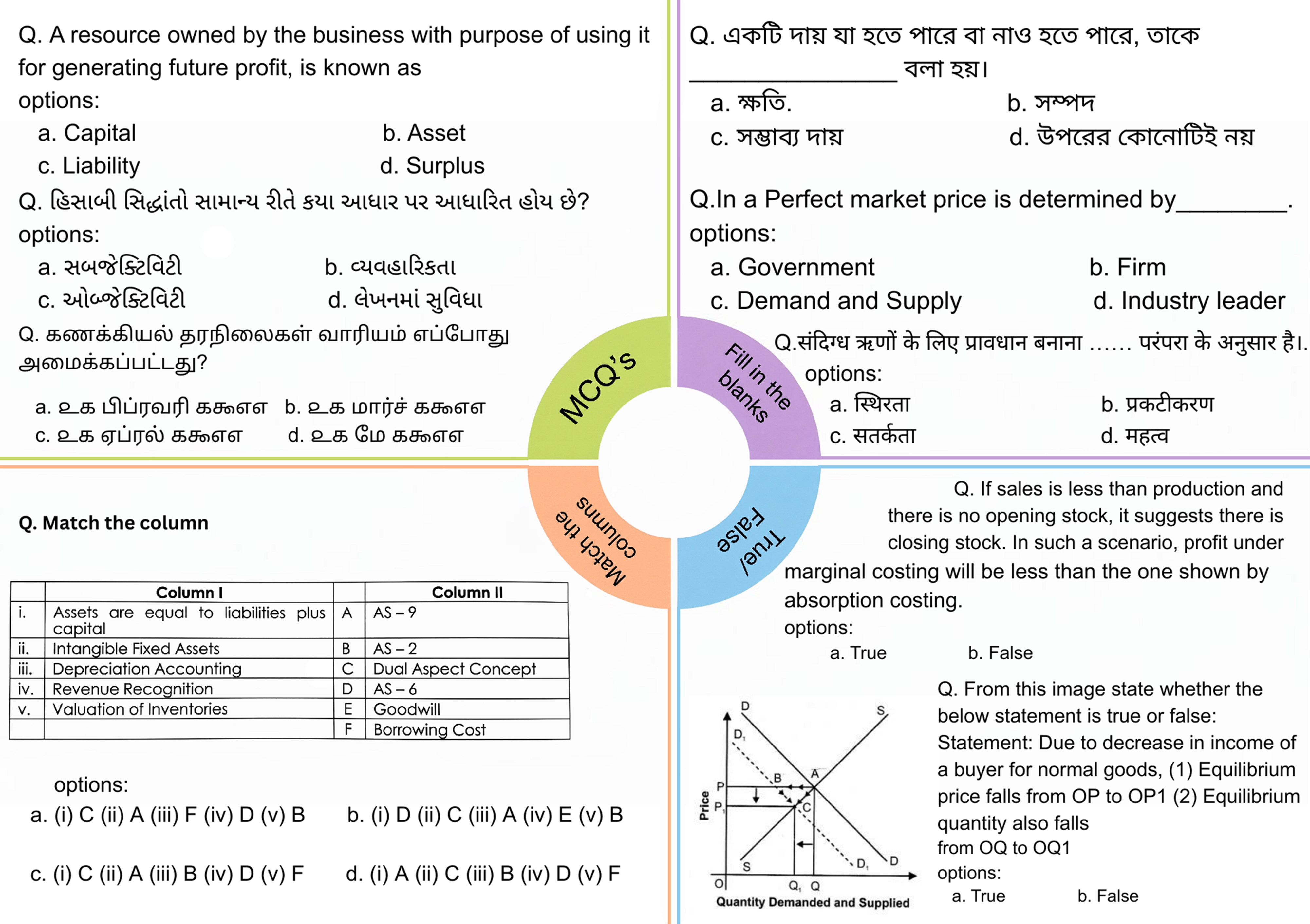}
    \caption{FinVQA Corpus Overview. Illustrative samples spanning four distinct question paradigms: multiple-choice (MCQ), fill-in-the-blanks, true/false, and table-matching.}
    \label{fig:overalldata}
\end{figure*}
However, real-world domain-specific numerical reasoning, particularly in finance, remains a significant challenge, as it requires precise mathematical computation, faithful application of domain knowledge, and robust reasoning over hybrid multimodal contexts such as tables, charts, and textual descriptions~\citep{chen2023theoremqa,das2023find,das2024investigate,romera2024mathematical,wang2024metacognitive}. Existing models often struggle with such hybrid reasoning, leading to errors in numerical accuracy and inconsistencies in multimodal grounding~\cite{tang2025finmmr,deng2025understanding}. Moreover, vision-language models (VLMs) tend to over-rely on textual cues while underutilizing visual financial signals, further limiting their effectiveness in realistic financial question answering scenarios~\cite{vo2025vision}.

These challenges are further amplified in multilingual settings such as India, where financial literacy initiatives and educational systems heavily rely on regional Indic languages. As the country undergoes rapid economic growth, enabling accurate and accessible financial reasoning across languages becomes increasingly critical. However, existing benchmarks largely focus on English and lack coverage of Indic languages, multimodal financial data, and controlled reasoning complexity.

To address this gap, we introduce FinVQA, a multilingual benchmark for financial visual question answering in Indic contexts. FinVQA comprises 18,900 samples spanning 14 financial domains, and is designed to capture diverse reasoning paradigms under realistic constraints. The dataset covers English, Hindi, Bengali, Marathi, Gujarati, and Tamil, and is organized into three difficulty levels (easy, moderate, hard) and four question formats: multiple choice, fill-in-the-blank, true/false, and table matching. Figure~\ref{fig:overalldata} illustrates representative examples from the dataset.

Building on this benchmark, we propose \textit{FIND}, a framework for reliable multimodal financial reasoning that combines supervised fine-tuning with constraint-aware decoding. FIND enforces structured outputs, improves numerical faithfulness, and promotes consistent multimodal grounding, enabling more trustworthy reasoning in high-stakes financial applications.

Our main contributions are as follows: 
\begin{itemize}
    \item We introduce \textit{FinVQA}, a large-scale benchmark for evaluating multilingual financial numerical with supported reasonings, comprising 18,900 samples across 14 financial domains, organized into three difficulty levels and four reasoning formats.
    \item Considering the linguistic diversity in India\footnote{\href{https://en.wikipedia.org/wiki/Languages_with_official_recognition_in_India}{Most Widely Spoken Languages in India}}, \textit{FinVQA} spans English and five major Indic languages, Hindi, Bengali, Marathi, Gujarati, and Tamil. The dataset is designed to evaluate realistic visual-textual financial reasoning.
    \item We propose \textit{FIND}, a constraint-aware framework that integrates supervised fine-tuning with structured decoding to improve numerical correctness, multimodal alignment, and reliability in financial reasoning tasks.
\end{itemize}

\begin{table*}
\centering
\caption{Comparison of multilingual financial question answering and related financial reasoning datasets}\label{dataset_comp}
\scriptsize
\tabcolsep2pt
\begin{tabular}{l|p{1in}|p{0.7in}|p{1in}|p{1.4in}|p{0.5in}|p{1in}}
\hline
& \textbf{Dataset Name} & \textbf{Count (approx.)} & \textbf{Type} & \textbf{Speciality} & \textbf{Modality} & \textbf{Language} \\ \hline
\multirow{25}{*}{\rotatebox{90}{\textbf{Non-Indic}}} & FinQA~\cite{chen2021finqa} & $\sim$8k QA pairs & Numerical QA & Multi-step numerical reasoning over financial reports & Text + Tables & English \\ \cline{2-7} 
 & TAT-QA~\cite{zhu2021tat} & $\sim$16.5k QA pairs & Table QA & Numerical reasoning over financial tables with text & Text + Tables & English \\ \cline{2-7} 
 & ConvFinQA~\cite{chen2022convfinqa} & $\sim$14.1k conv. turns & Conversational QA & Multi-turn numerical reasoning over financial data & Text + Tables & English \\ \cline{2-7} 
 & MultiHiertt~\cite{klimaszewski2025avenibench} & $\sim$10k sentence pairs & NLI / reasoning & Hierarchical text-table entailment, includes finance & Text + Tables & English \\ \cline{2-7} 
 & FinMME~\cite{luo2025finmme} & $\sim$11k samples & Multimodal QA & Financial chart/table understanding and reasoning & Text + Charts + Tables & English \\ \cline{2-7} 
 & FinDER~\cite{choi2025finder} & $\sim$5.7k queries & Retrieval QA & Retrieval-augmented financial QA & Text & English \\ \cline{2-7} 
 & FinLLMs~\cite{yuan2024finllms} (core dataset) & $\sim$15k synthetic QA & Numerical QA & Programmatic generation of financial reasoning problems & Text + Tables & English \\ \cline{2-7} 
 & FinTruthQA~\cite{xu2024fintruthqa} & $\sim$6k QA pairs & Investor-firms interactions & Investor questions and Company responses & Text & Chinese\\ \cline{1-7} 
\multirow{8}{*}{\rotatebox{90}{\textbf{Indic}}} & IndicFinNLP (Exaggerated Numeral   Detection)~\cite{ghosh2024indicfinnlp} & $\sim$6.5k financial statements & Binary classification & Sustainability Classification in financial texts & Text & Hindi, Bengali, Telugu \\ \cline{2-7} 
 & Proposed \textbf{\textit{FinVQA}} & 18,900 & MCQ, Fill in the blanks, True \& False, Column Matching & Financial Problem specific reasonings & Text + Images & English, Hindi, Bengali, Gujarati, Marathi, Tamil \\ \hline
\end{tabular}
\end{table*}

\section{Background}
Recent progress in large language models equipped with explicit deliberative and multi-step inference mechanisms, such as OpenAI o1 \cite{2024d} and DeepSeek-R1 \cite{2025guo}, has significantly improved multi-step reasoning performance across domains, including code~\cite{chen2021evaluating,jain2024livecodebench}, mathematics~\cite{lightman2023let,mao2024champ}, and science~\cite{lu2023mathvista,wang2023scibench,yue2024mmmu}. However, their capabilities on domain-specific numerical reasoning, particularly in finance, remain insufficiently validated.

Existing financial tasks~\citep{das2024negative,das2025deciphering} and their reasoning benchmarks, such as FinQA \cite{chen2021finqa}, TAT-QA \cite{zhu2021tat}, ConvFinQA, and their programmatic extensions CodeFinQA and CodeTAT-QA \cite{krumdick2024bizbench} largely focus on tabular extraction and shallow arithmetic, limiting their ability to distinguish advanced LRMs from standard LLMs. Other datasets, including FinCode \cite{krumdick2024bizbench} and FinanceMath \cite{zhao2024financemath}, suffer from limited complexity, ambiguous problem formulations, and relaxed evaluation protocols, hindering accurate assessment of true reasoning ability.

Moreover, as evidenced by existing benchmarks (Table~\ref{dataset_comp}), financial reasoning datasets are overwhelmingly English-centric, with minimal support for multilingual or Indic-language settings, and limited coverage of multimodal financial VQA. This gap is particularly critical in regions such as India, where financial education and decision-making often rely on regional languages. These limitations highlight the need for a multilingual, multimodal, and rigorously annotated financial reasoning benchmark that evaluates complex, domain-aware reasoning beyond surface-level numerical accuracy motivating the development of \textit{FIND}\footnote{\href{https://github.com/sarmistha-D/FIND_FinVQA}{Resources for \textit{FIND} and \textit{FinVQA}}}.

\section{FinVQA Dataset Curation}

To promote societal impact through education-oriented artificial intelligence, we curate a high-quality collection of \emph{financial and economic reasoning questions} from authoritative Indian educational sources. Our primary data source is textbooks published by the National Council of Educational Research and Training (NCERT), India’s apex body for school education, covering core financial concepts in \emph{Accountancy, Business Studies, and Economics} for Classes 9--12\footnote{\href{https://www.learncbse.in/ncert-solutions-class-12-micro-economics-demand/}{NCERT website}}. These materials provide structured coverage of topics such as financial accounting, financial management, and demand supply analysis.
\begin{figure}[!h]
    \centering
    \includegraphics[width=\columnwidth]{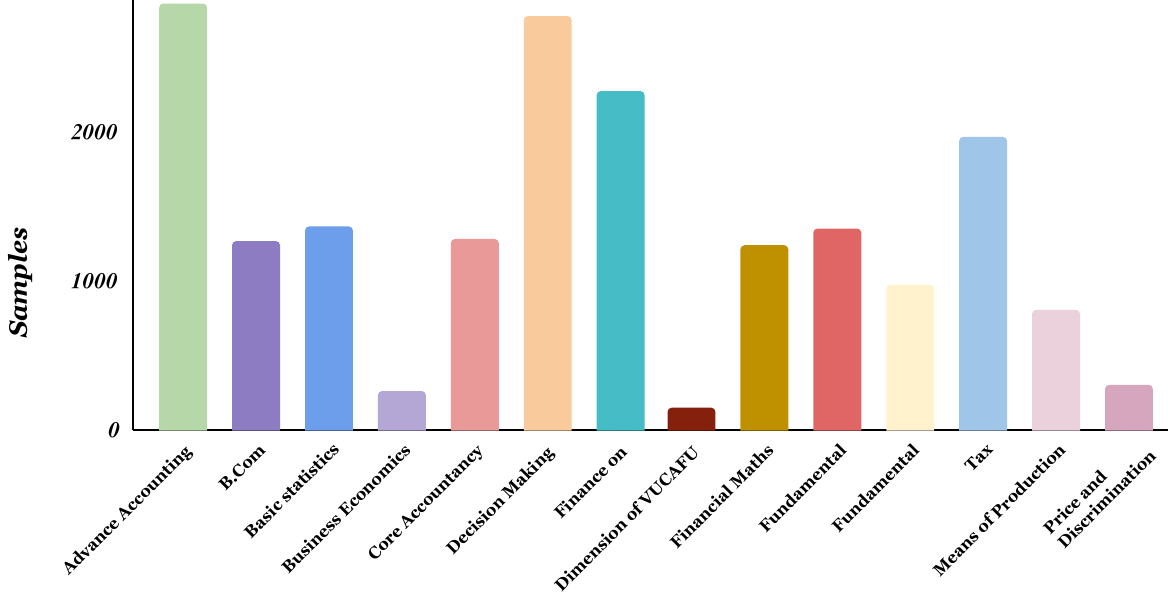}
    \caption{Domain-wise distribution of FinVQA.}
    \label{fig:barplot_half}
\end{figure}
\begin{figure}[!h]
    \centering
    \includegraphics[width=0.7\columnwidth]{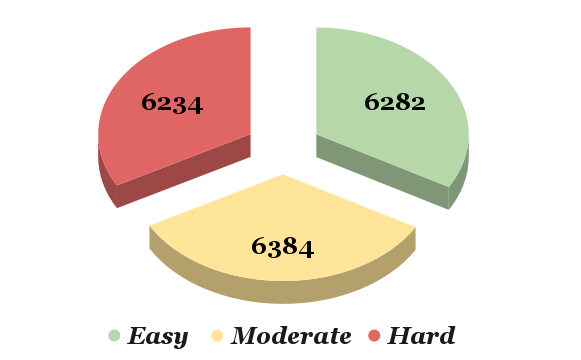}
    \caption{Difficulty-wise distribution of FinVQA.}
    \label{fig:pieplot_half}
\end{figure}
To complement this foundation, we additionally incorporate professionally oriented content from the ICMAI--CMA program, administered by the \emph{Institute of Cost Accountants of India (ICMAI)}, a statutory body under the Ministry of Corporate Affairs\footnote{\href{https://icmai.in/studentswebsite/Syl-2022-Fdn-Stdy-Mtrls.php}{ICMAI website}}. From these curated sources, we initially collect over 3,000 English-language samples spanning both \emph{unimodal textual questions} and \emph{multimodal visual question answering (VQA)} formats. Following rigorous manual scrutiny and quality control, we retain 2519 text-only samples and 631 image-based financial reasoning samples. This filtering process ensures unambiguous problem formulation, numerical correctness, and suitability for evaluating both conceptual understanding and quantitative reasoning in financial contexts.\\
\subsection{Multilingual Data Curation} 

To construct the multilingual textual samples, we employed \textsc{GPT-4o} to translate the original English instances into five target languages: Hindi, Bengali, Gujarati, Marathi, and Tamil (see Appendix~\ref{label:dataset_annotation} for further details). Translation quality was quantitatively evaluated using cosine similarity over sentence embeddings generated by the \textit{multilingual-MiniLM-L12-v2}~\cite{reimers-2019-sentence-bert} Sentence Transformer model, thereby ensuring semantic alignment across languages. To further strengthen translation reliability, a back-translation procedure was additionally performed. Finally, for authoritative validation, we engaged five native speakers one per target language who manually reviewed each translated instance and resolved linguistic ambiguities, idiomatic inconsistencies, and contextual deviations, ensuring high linguistic accuracy and cultural fidelity of the multilingual corpus.

\subsection{OCR-Based Text Replacement} 
Images in the dataset contain English text which also needs to be translated to target languages. We utilized PaddleOCR~\cite{cui2025paddleocrvlboostingmultilingualdocument} that outputs structured information for each detected text string from these images, including the recognized text, its bounding-box coordinates, and an orientation flag indicating horizontal or vertical alignment. Leveraging this metadata, the original English text is replaced with its translated counterpart at the exact spatial location from which it was extracted. Figure~\ref{fig:ocr_based_text} in the appendix represents one instance. 

To maintain visual coherence, the original text region is first removed via background-matched inpainting, followed by rendering the translated text within the same bounding box, thereby preserving the original layout and spatial structure.

For rendering Indic-language translations, we employ the \texttt{NotoSans-\{IndicLanguage\}} font families to ensure consistent glyph coverage and script fidelity. The font size is adaptively determined based on the detected region geometry; specifically, for horizontally aligned text, the font size is set to 90\% of the bounding-box height, ensuring optimal readability while respecting spatial constraints.

\subsection{Back-Translation} 
We employ back-translation to enhance linguistic diversity and robustness in multilingual settings by translating instances to an intermediate language and back, preserving semantics. For \textit{FIND}, back-translations are generated using GPT-4o, and semantic fidelity is verified via cosine similarity, with most instances achieving 85–90\% similarity to the originals. Low-similarity cases are manually reviewed to ensure numerical consistency and constraint preservation.

\subsection{Dataset Annotation and Quality Check}

\begin{figure*}[t]
    \centering
    
    \begin{subfigure}[t]{0.48\textwidth}
        \centering
        \includegraphics[width=\linewidth]{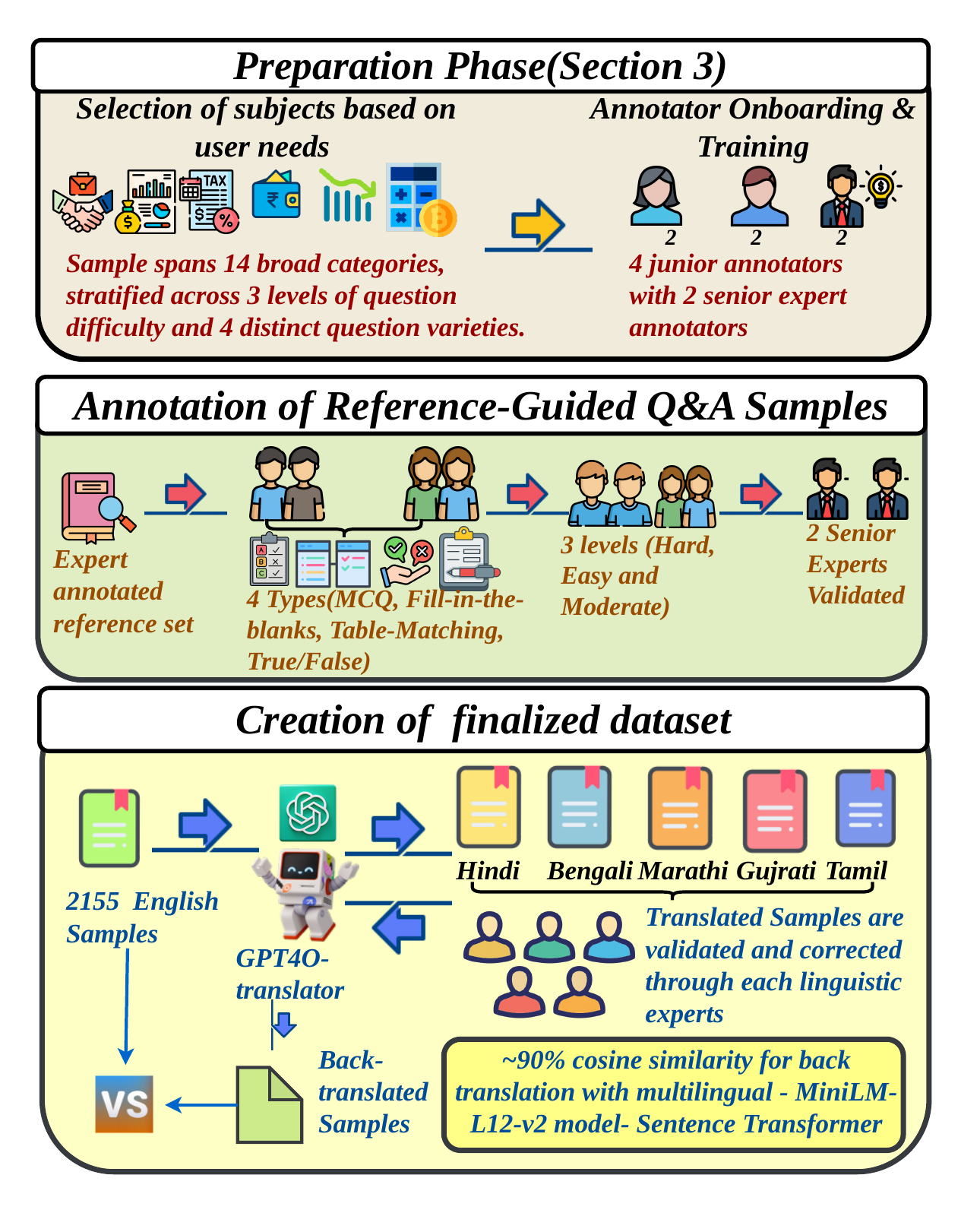}
        \caption{End-to-End Dataset Annotation and Construction Pipeline.}
        \label{fig:annotation}
    \end{subfigure}
    \hfill
    \begin{subfigure}[t]{0.48\textwidth}
        \centering
        \includegraphics[width=\linewidth]{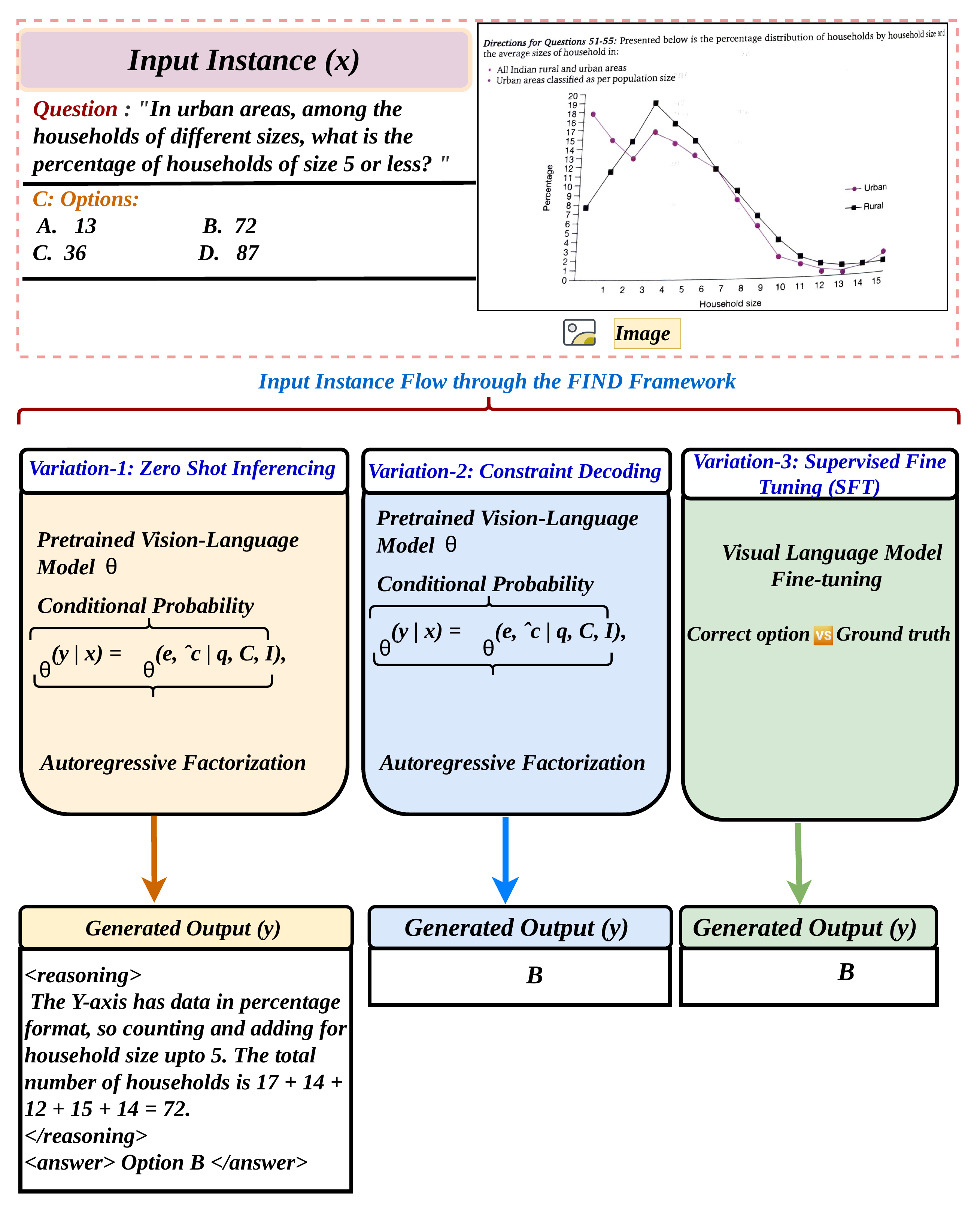}
        \caption{High-level architecture of the proposed \textit{FIND} framework for multimodal financial QA.}
        \label{fig:architecture}
    \end{subfigure}
    
    \caption{Overview of the \textit{FinVQA} dataset construction pipeline and the proposed \textit{FIND }framework.}
    \label{fig:combined}
\end{figure*}

To construct \textit{FinVQA}, we adopted a structured human annotation pipeline as represented in Figure~\ref{fig:annotation} involving six annotators. Among them, four junior annotators were undergraduate-level students, while the remaining two senior annotators were domain experts with professional experience in finance. To establish consistent annotation standards, we initially used 3150 English data samples before translation, and senior experts first annotated 200 seed samples, comprising both image-based and text-based questions, which served as reference exemplars for training junior annotators. Detailed annotation guidelines are provided in the Appendix~\ref{label:dataset_annotation}.

During the final annotation phase, all junior annotators were required to strictly adhere to the predefined annotation rules. Upon completion, the senior experts conducted an \emph{inter-annotator agreement} analysis, achieving a Fleiss' Kappa score~\cite{gwet2014handbook} of 0.78, indicating substantial agreement. The dataset was subsequently refined through expert-led validation based on the following criteria:

\begin{enumerate}
    \item \textbf{Question Framing and Option Balance:} Each question was examined to ensure clear formulation and an equitable distribution of multiple-choice options (a, b, c, d). For numerical problems, structured and logically consistent reasoning traces were verified and retained.
    \item \noindent \textbf{Difficulty and Question-Type Distribution:} The dataset was balanced across \emph{easy}, \emph{moderate}, and \emph{hard} difficulty levels, while maintaining diversity in question formats, including fill-in-the-blank, true/false, and table-matching tasks.
    \item \noindent \textbf{Expert Correction and Multimodal Validation:} Any detected error in questions, answer options, numerical values, or reasoning steps were manually corrected by  senior annotators. For multimodal samples, image quality and semantic alignment with the corresponding textual context were rigorously verified to ensure clarity and relevance.
\end{enumerate}

Conclusively, the proposed \textit{FinVQA} corpus comprises a total of 18,900 carefully curated instances, including 15,114 text-only and 3,786 image-based samples. The dataset spans 14 distinct financial domains and is systematically organized across three levels of question difficulty: easy (6282), moderate (6384), and hard (6234). The domain-wise distribution and difficulty-wise distribution are illustrated in Figures~\ref{fig:barplot_half} and~\ref{fig:pieplot_half} respectively, while comprehensive dataset statistics and construction details are provided in Appendix~\ref{sec:data}.

\section{The \textit{FIND} Methodology}
\subsection{Multimodal Financial QA Problem}

We address multimodal financial reasoning using a multiple‐choice question answering (MCQA) formulation. Each instance in the dataset $\mathcal{D}$ comprises a question $q$, a set of candidate answers $\mathcal{C}=\{c_1,\ldots,c_K\}$, optional visual input $\mathcal{I}$ (e.g., chart, table, receipt, or document), the correct answer $c^*\in\mathcal{C}$, and a ground‐truth explanation represented as \(\{q,\mathcal{C},\mathcal{I},c^*\}\sim\mathcal{D}\). The learning objective must encourage solutions that are both accurate and logically coherent with their explanations, reflecting the reasoning demands of financial tasks. Our proposed framework is illustrated in Figure~\ref{fig:architecture}.


\subsection{The \textit{FIND} Learning Framework}
To systematically study multimodal financial reasoning under varying degrees of supervision and output control, we design the experimental pipeline with three-variations, namely : (i) zero-shot inference, (ii) constrained decoding, and (iii) supervised fine-tuning (SFT). Each phase progressively increases task specialization and structural enforcement while operating over the same underlying MCQA formulation.

\noindent Let $\pi_{\theta}$ denote a vision-language model parameterized by $\theta$, which defines a conditional distribution over output sequences given multimodal inputs. To conduct these tasks the subsequent three variant experimental pipelines are discussed below. The corresponding prompts are given in Appendix \ref{label:prompts}.

\subsubsection{Variation I: Zero-Shot Inference}

In the first phase, we evaluate the intrinsic reasoning capability of pretrained vision--language models without task-specific adaptation. Given an input instance \( x = \{q, \mathcal{C}, I\} \), the model directly estimates the conditional probability \( \pi_{\theta}(y \mid x) = \pi_{\theta}(\hat{e}, \hat{c} \mid q, \mathcal{C}, I) \), where \( \hat{e} \) denotes the generated reasoning sequence and \( \hat{c} \in \mathcal{C} \) is the predicted answer option. Inference is performed autoregressively by factorizing the output distribution as \( \pi_{\theta}(y \mid x) = \prod_{t=1}^{T} \pi_{\theta}(y_t \mid y_{<t}, q, \mathcal{C}, I) \), without explicit constraints on output structure or length. While this setup enables unbiased evaluation of generalization, it often results in formatting instability, verbosity, and language inconsistency, particularly in multilingual financial settings.

\subsubsection{Variation II: Constrained Decoding for Structured Prediction}\label{decoding}

To mitigate the common failure modes such as overly verbose responses, format violations, code-switching across languages, and the generation of repetitive or irrelevant content,  observed in zero-shot inference we introduce constrained decoding at inference time implemented using \textsc{vLLM}~\cite{kwon2023efficient}. Technically, this is realized by restricting the model’s output space to valid multiple-choice answer option tokens (e.g., A/B/C/D) while preserving the internal reasoning process.

Let $\mathcal{V}$ denote the full tokenizer vocabulary and $\mathcal{V}_{\text{valid}} \subset \mathcal{V}$ the subset corresponding to valid answer options. During decoding, the token-level distribution is constrained as 
$ \tilde{\pi}_{\theta}(y_t\mid\cdot) = \pi_{\theta}(y_t \mid \cdot)\ \text{if}\ y_t \in \mathcal{V}_{\text{valid}},\ \text{and}\ 0\ \text{otherwise}$,
implemented by masking invalid logits with $-\infty$ prior to softmax. The decoding objective is thus 
\( \hat{c} = \arg\max_{c_k \in \mathcal{C}} \sum_{t=1}^{|c_k|} \log \pi_{\theta}(c_{k,t} \mid c_{k,<t}, x) \),
where $|c_k|$ denotes the number of subword tokens in option $c_k$.   To account for tokenizer variability, we apply language-aware constraints: single-token decoding for English, Hindi, and Marathi, and up to three tokens for Bengali, Gujarati, and Tamil. This strategy enforces strict format compliance, suppresses spurious generations, and ensures stable, comparable predictions across languages without modifying model parameters.

\subsubsection{Variation III: SFT of Vision Language Models}


In this variant, we apply SFT to align the vision--language model with the financial \textbf{MCQA} task under answer-only supervision. Given a labeled dataset \( \mathcal{D} = \{(x_i, c_i^*)\}_{i=1}^{N} \), where \( c_i^* \in \mathcal{C}_i \) denotes the correct option for input \( x_i \), the model is trained to generate the target answer text corresponding to the correct choice.

Since answer options may consist of multiple tokens, the model learns to predict the sequence of tokens forming the correct answer conditioned on the input. The training objective maximizes the likelihood of generating this correct answer sequence given the input.

\noindent \textbf{Parameter-Efficient Attention Adaptation.}
To limit overfitting and training cost, we fine-tune only the query, key, and value projections in self- and cross-attention layers. For each layer, projections are computed as \( Q = XW_Q,\ K = XW_K,\ V = XW_V \), and optimization is restricted to \( \theta_{\text{train}} = \{W_Q^{(\ell)}, W_K^{(\ell)}, W_V^{(\ell)}\}_{\ell=1}^{L} \), while all remaining parameters \( \theta_{\text{frozen}} = \theta \setminus \theta_{\text{train}} \) are held fixed. Attention is computed as \( \mathrm{Attn}(Q,K,V) = \mathrm{softmax}\!\left(\frac{QK^\top}{\sqrt{d_k}}\right)V \). To further enhance parameter efficiency, we utilise Low-Rank Adaptation (LoRA)~\cite{hu2022lora} with rank \( r = 8 \), scaling factor \( \alpha = 32 \), and dropout rate of 0.1.

This design enables selective recalibration of multimodal interactions, particularly between visual representations \( \mathcal{I} \) and textual queries, while mitigating catastrophic forgetting. Although reasoning tokens are not verbalised or explicitly supervised, latent reasoning patterns emerge through attention reweighting induced by answer-level supervision. To prevent verbosity and format violations, the model is trained with SFT and deployed with the constrained decoding strategy from variation~II, ensuring high predictive accuracy and strict structural validity.

The dataset was split into training and testing sets in the ratio 70:30. For all variations the metrics are reported using the test set. We train models for 3 epochs, with the learning rate for training fixed at 1e-5, and the optimiser used was the 8-bit variant of AdamW. All of the experiments are conducted on a machine with one Nvidia A100 80GB GPU.

\section{Experiments and Results}
This section presents the experimental setup, research questions, and the principal conclusions drawn from the empirical evaluation.
Our study is guided by the following research questions:
\begin{itemize}
    \item \textbf{RQ1:} What is the overall performance of the evaluated models? 
    \item \textbf{RQ2:} Which modeling technique exerts the greatest influence on performance?
    \item \textbf{RQ3:} What language-specific performance trends and insights emerge?
    \item \textbf{RQ4:} How do the experimental results justify the chosen model framework design?
\end{itemize}

\noindent\textbf{Models and Evaluation Protocol.}  
We evaluate a diverse set of state-of-the-art VLMs spanning multiple parameter scales, including compact (4B), medium (7B–12B), and large-scale (>27B) architectures. Specifically, our experiments consider models from the \textit{Qwen2.5-VL}~\cite{qwen25vltechnicalreport}, \textit{Qwen3-VL}~\cite{qwen3technicalreport} and \textit{Gemma-3}~\cite{gemma_2025} families. Each model is assessed under zero-shot, constrained-decoding, and SFT settings. Model performance is quantitatively evaluated using accuracy as the primary metric.

\subsection{Analytical Discussion}

\begin{figure*}[h]
    \centering
    \begin{subfigure}{\columnwidth}
        \centering
        \includegraphics[width=\linewidth]{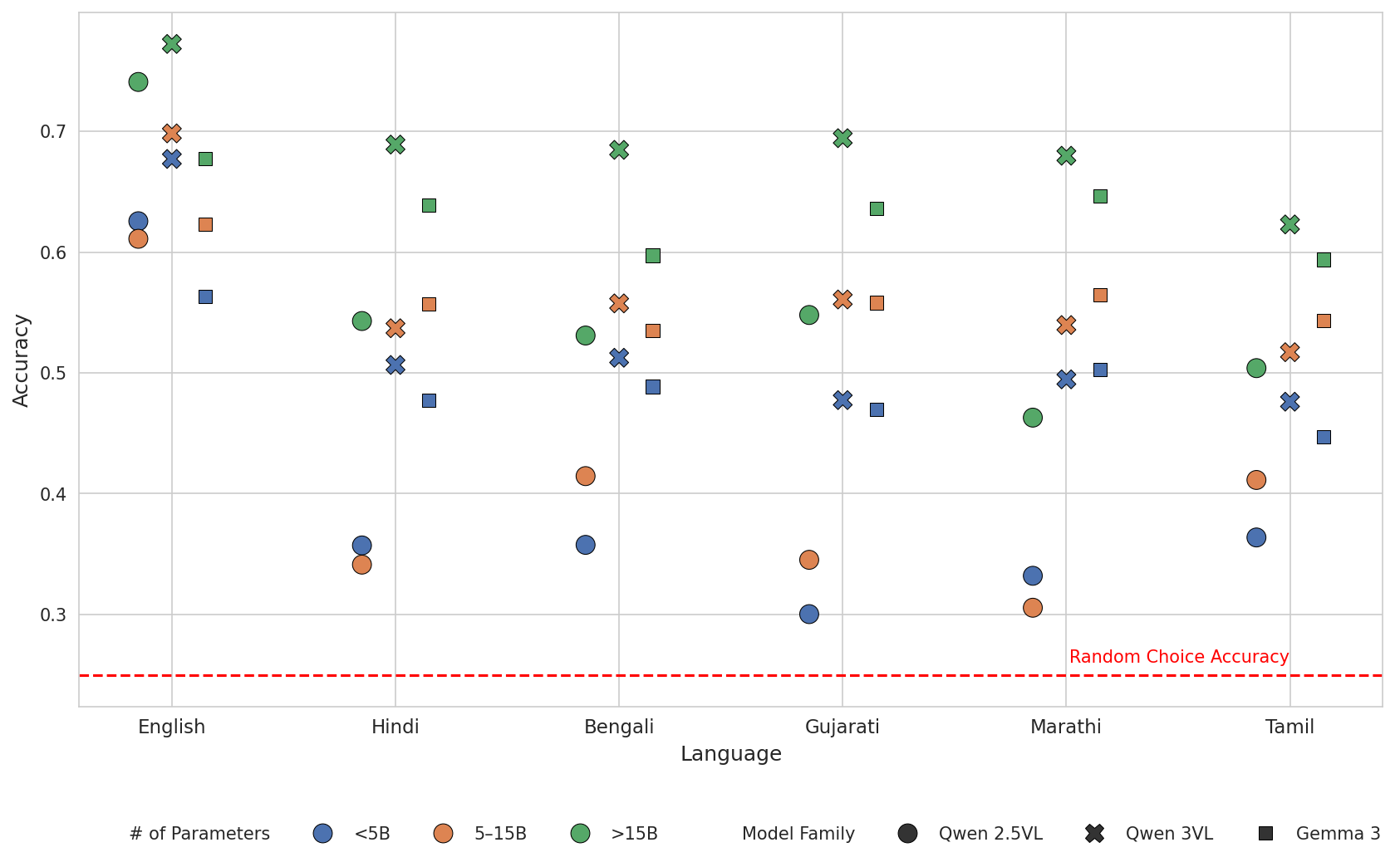}
        \caption{Text-only dataset}
        \label{fig:sft3-text}
    \end{subfigure}\hfill
    \begin{subfigure}{\columnwidth}
        \centering
        \includegraphics[width=\linewidth]{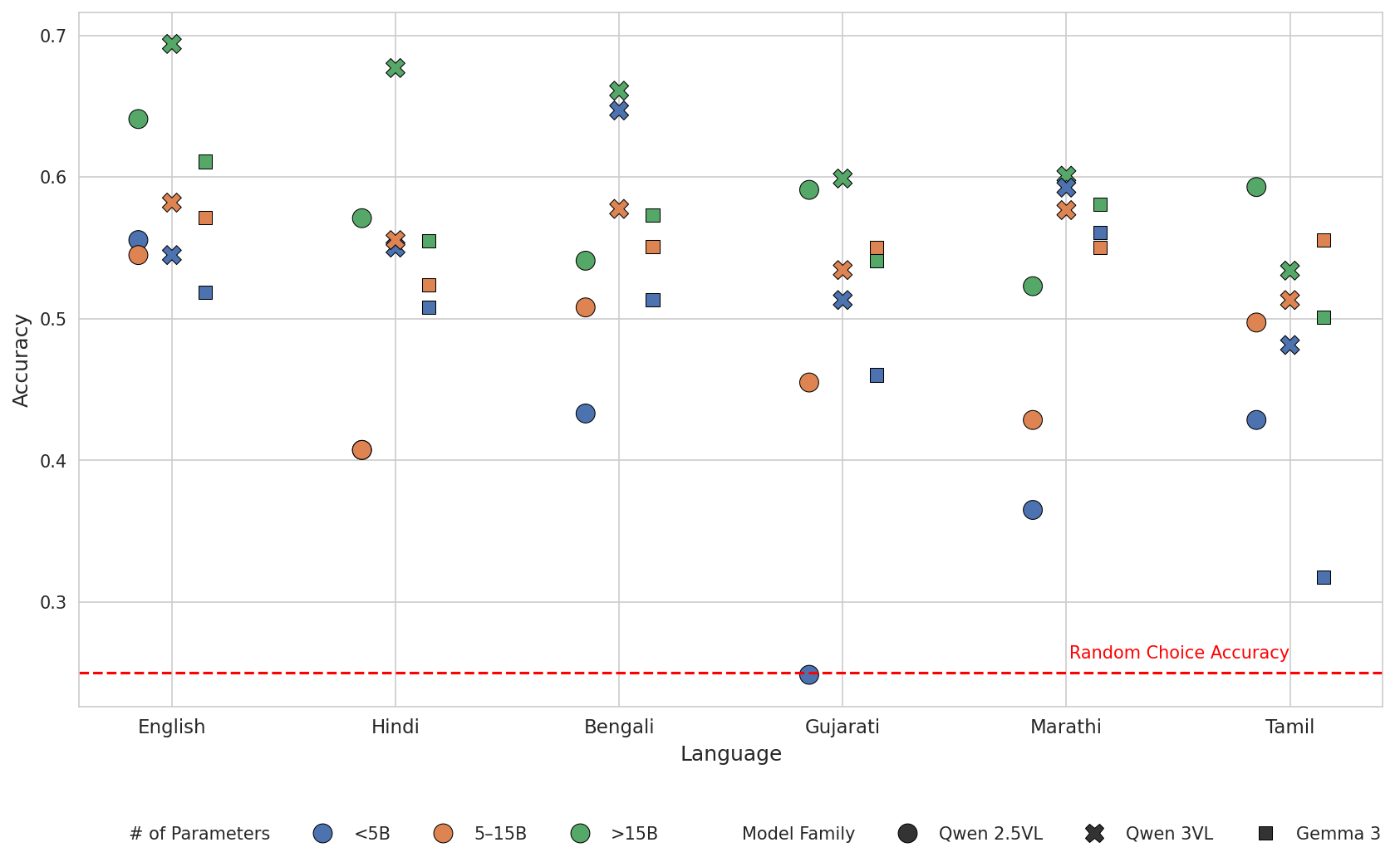}
        \caption{Text + image dataset}
        \label{fig:sft3-multimodal}
    \end{subfigure}

    \caption{Inference results of the SFT models.}
    \label{fig:sft3-results}
\end{figure*}

\begin{table*}
\centering
\tabcolsep4pt
\caption{Model-wise Accuracy for text only part of dataset across Indic languages using zero shot inference, constrained decoding, and SFT models inferenced with constrained decoding}
\label{tab:text_results}
\scriptsize
\begin{tabular}{lcccccccccccccccccc}
\hline
\multirow{2}{*}{\textbf{Model Name}} & 
\multicolumn{6}{c}{\textbf{Zero shot inference}} &\multicolumn{6}{c}{\textbf{Constrained decoding}} &\multicolumn{6}{c}{\textbf{Supervised finetuning}} \\
\cline{2-19}
 & \textbf{en} & \textbf{hi} & \textbf{bn} & \textbf{mr} & \textbf{gu} & \textbf{ta}& \textbf{en} & \textbf{hi} & \textbf{bn} & \textbf{mr} & \textbf{gu} & \textbf{ta}& \textbf{en} & \textbf{hi} & \textbf{bn} & \textbf{mr} & \textbf{gu} & \textbf{ta} \\
\hline
Qwen2.5-VL-3B-Instruct  &57.5&26.7&27.3&14.3&25.4&21.2&63.1&36.9&20.9&29.4&32.4&20.2&62.6&35.7&35.8&30.0&33.2&36.4\\
Qwen3-VL-4B-Instruct    &69.6&51.2&48.5&26.2&39.6&17.5&67.2&49.6&40.8&14.3&49.1&46.7&67.7&50.7&51.3&47.8&49.5&47.6\\
Gemma-3-4b-it           &61.5&50.3&52.2&50.4&41.3&45.2&56.3&46.3&49.1&48.4&49.6&40.9&56.3&47.8&48.9&47.0&50.3&44.7\\
\hline
Qwen2.5-VL-7B-Instruct  &55.6&36.6&37.4&20.5&31.3&27.6&61.2&39.8&34.2&18.3&34.1&40.1&61.1&34.1&41.5&34.5&30.6&41.1\\
Qwen3-VL-8B-Instruct    &74.1&58.5&58.4&49.5&56.5&46.4&69.0&53.4&45.0&28.0&53.6&52.1&69.8&53.7&55.8&56.1&54.0&51.7\\
Gemma-3-12b-it          &71.7&62.7&66.6&63.4&64.3&57.8&62.3&58.2&58.5&57.3&59.3&55.3&62.3&55.7&53.5&55.8&56.5&54.4\\
\hline
Qwen2.5-VL-32B-Instruct &73.8&40.6&27.9&22.6&34.4&33.9&70.5&48.9&35.1&35.4&45.0&50.4&74.1&54.3&53.1&54.8&46.3&50.4\\
Qwen3-VL-32B-Instruct   &77.8&70.9&66.1&60.6&69.4&57.3&76.9&68.9&52.1&33.6&67.2&61.2&77.2&68.9&68.5&69.4&68.0&62.3\\
Gemma-3-27b-it          &73.7&70.4&67.4&69.7&68.9&65.6&69.2&64.0&61.3&63.4&64.3&61.2&67.7&63.9&59.7&63.6&64.7&59.4\\
\hline
\end{tabular}
\end{table*}

\begin{table*}
\centering
\tabcolsep4pt
\caption{Model-wise Accuracy for multimodal version of dataset across Indic languages using zero shot inference, constrained decoding, and SFT models inferenced with constrained decoding}
\label{tab:multimodal_results}
\scriptsize
\begin{tabular}{lcccccccccccccccccc}
\hline
\multirow{2}{*}{\textbf{Model Name}} & 
\multicolumn{6}{c}{\textbf{Zero shot inference}} &\multicolumn{6}{c}{\textbf{Constrained decoding}} &\multicolumn{6}{c}{\textbf{Supervised finetuning}} \\
\cline{2-19}
 & \textbf{en} & \textbf{hi} & \textbf{bn} & \textbf{mr} & \textbf{gu} & \textbf{ta}& \textbf{en} & \textbf{hi} & \textbf{bn} & \textbf{mr} & \textbf{gu} & \textbf{ta}& \textbf{en} & \textbf{hi} & \textbf{bn} & \textbf{mr} & \textbf{gu} & \textbf{ta} \\
\hline
Qwen2.5-VL-3B-Instruct&52.4&33.9&31.0&13.2&25.9&22.2&54.5&41.8&46.5&24.3&35.4&40.2&55.6&40.7&43.3&24.9&36.5&42.9\\
Qwen3-VL-4B-Instruct&61.9&46.0&47.1&19.0&37.0&12.2&55.6&55.0&62.0&50.8&60.8&46.0&54.5&55.0&64.7&51.3&59.3&48.1\\
Gemma-3-4b-it&37.6&44.4&42.8&41.8&43.4&30.2&53.4&52.9&54.5&44.4&54.0&29.6&51.9&50.8&51.3&46.0&56.1&31.7\\
\hline
Qwen2.5-VL-7B-Instruct&61.4&46.6&49.2&26.5&42.9&45.5&53.4&41.3&52.4&47.6&47.6&49.2&54.5&40.7&50.8&45.5&42.9&49.7\\
Qwen3-VL-8B-Instruct&64.6&49.2&47.6&27.5&44.4&41.8&58.2&55.0&57.8&52.9&56.6&51.3&58.2&55.6&57.8&53.4&57.7&51.3\\
Gemma-3-4b-it&66.7&55.0&60.4&58.7&60.3&58.2&59.3&52.4&53.5&52.4&54.5&54.5&57.1&52.4&55.1&55.0&55.0&55.6\\
\hline
Qwen2.5-VL-32B-Instruct&67.2&36.0&22.5&9.0&32.8&30.1&58.7&55.6&54.5&54.5&48.1&55.0&64.1&57.1&54.1&59.1&52.3&59.3\\
Qwen3-VL-32B-Instruct&68.3&51.8&49.1&31.3&49.1&45.9&62.4&63.0&65.2&58.7&61.9&53.4&69.4&67.7&66.1&59.9&60.1&53.4\\
Gemma-3-4b-it&69.3&62.4&59.9&55.6&61.9&63.5&59.3&52.9&58.8&52.9&59.8&50.8&61.1&55.5&57.3&54.1&58.1&50.1\\
\hline
\end{tabular}
\end{table*}

This section synthesises the empirical findings and distils key insights in relation to the research questions.
\subsubsection{Answer to RQ1 (Overall Performance)} 
Across both text-only and multimodal settings, model performance exhibits a clear hierarchy governed by model scale and training regime. In the zero-shot text setting (Table \ref{tab:text_results}), smaller models ($\leq$4B) show weak multilingual generalization, with accuracies often below 30 for Marathi, Gujarati, and Tamil (e.g., Qwen2.5-VL-3B: 14.3 in Marathi, 21.2 in Tamil; Qwen3-VL-4B: 17.5 in Tamil). In contrast, large-scale models perform substantially better; for instance, Qwen3-VL-32B-Instruct achieves 70.9 (Hindi), 66.1 (Bengali), and 60.6 (Marathi) in zero-shot inference. Following supervised fine-tuning, text-only performance improves considerably, even with Qwen3-VL-32B-Instruct reaching balanced accuracies across Indic languages such as 68.9 (Hindi), 68.5 (Bengali), 69.4 (Marathi), and 68.0 (Gujarati) while smaller models (3B–7B) remain below 40 in low-resource languages. In multimodal evaluations, constrained decoding and SFT (Table \ref{tab:multimodal_results}) progressively enhance performance over zero-shot baselines, particularly for Indic languages. Overall, English consistently attains the highest accuracy, and large-scale VLMs outperform smaller counterparts across all modalities and evaluation regimes, confirming strong capacity-dependent cross-lingual generalization.

\subsubsection{Answer to RQ2 (Greatest Influence)}
The results show that SFT exerts the strongest influence on performance, substantially outweighing gains from constrained decoding alone. While constrained decoding yields modest and sometimes inconsistent improvements, SFT produces pronounced cross-lingual gains. For example, Qwen3-VL-4B improves Marathi accuracy from 26.2 (zero-shot) to 47.8 (+21.6 points) after SFT, with similar trends in Gujarati and Tamil. 

In the multimodal setting, Qwen3-VL-8B increases Marathi accuracy from 27.5 (zero-shot) to 52.9 with constrained decoding, and further stabilizes around 53.4 after SFT. Likewise, Qwen3-VL-32B improves Bengali accuracy from 49.1 (zero-shot) to 65.2 (constrained decoding) and reaches 66.1 post-SFT. Overall, while constrained decoding aids structural validity, SFT delivers the most consistent and robust multilingual improvements, with additional epochs yielding diminishing but stabilizing gains.

\subsubsection{Answer to RQ3 (Linguistic trends)}
Language-wise analysis reveals stable and interpretable trends across modalities and training regimes. Hindi and Bengali consistently achieve the highest accuracies, often exceeding 60 in large models, reflecting higher resource availability and closer alignment with English. Marathi and Gujarati exhibit the largest relative gains from constrained decoding and SFT, frequently improving by 15-30 absolute points compared to zero-shot multimodal inference (e.g., Qwen3-VL-4B Marathi: 19.0 → 50.8 → 51.3), highlighting their sensitivity to alignment-based training. 

Tamil remains the most challenging language, with lower absolute accuracies (below 65 even after SFT), likely due to greater typological distance and script complexity; nevertheless, it shows consistent post-SFT improvements. Overall, SFT substantially reduces cross-lingual performance variance, indicating enhanced language-agnostic reasoning rather than superficial transfer.

\subsubsection{Answer to RQ4 (Framework Justification)} 
The empirical results strongly validate the proposed framework, which combines large-scale instruction-tuned VLMs with supervised alignment and constrained decoding. Large models such as Qwen3-VL-32B and Gemma-3-27B consistently achieve the highest and most stable performance across text-only and multimodal settings (Figure \ref{fig:sft3-results}), often yielding 15-30 point gains over zero-shot baselines in low-resource Indic languages after SFT. 

Smaller models become competitive only after supervised alignment, underscoring the importance of capacity-aware architectures. In contrast, weakly aligned models (e.g., LLaVA variants) fail under constrained decoding, exhibiting near-zero accuracies across Indic languages. Additionally, the diminishing yet stabilizing gains across successive SFT epochs confirm that early alignment combined with controlled inference rather than reliance on zero-shot generalization is critical for robust multilingual multimodal reasoning, thereby justifying the framework’s design choices.

\subsection{Human Evaluation} 

We conduct a comprehensive human evaluation of the zero-shot inference results adhering to a well-defined evaluation protocol in appendix \ref{human_eval_guidelines}. The assessment is grounded in a structured rubric designed to capture critical dimensions of financial question answering quality, including Financial Domain Understanding, Problem Interpretation, Mathematical Correctness, Formula Application, Reasoning Consistency, and Formatting Compliance, enabling fine-grained assessment of reasoning and usability.

The results (Table \ref{tab:financial_reasoning_human_eval} in the appendix) show a strong scale–performance relationship. While Formatting Compliance remains relatively stable across models ($\approx$0.7–0.8), higher-order competencies particularly Financial Domain Understanding and Problem Interpretation are strongly capacity-dependent. 
Smaller models (e.g., Qwen2.5-VL-3B, Gemma-3-4B) score as low as 0.1-0.2, whereas large models (Gemma-3-27B, Qwen3-VL-32B) approach near-ceiling performance (0.9-1.0). The Qwen3-VL family consistently outperforms comparable baselines, with Qwen3-VL-32B achieving the highest aggregate score (0.967). Overall, syntactic compliance is achievable at low compute, but high-fidelity financial reasoning requires large-scale models.

\subsection{Qualitative Analysis}

A qualitative analysis of the models across three parameter scales reveals a distinct progression in multimodal and mathematical reasoning capabilities. As showcased using a representative sample in Table \ref{tab:qualitative_sample} in the appendix, at the 3-4B parameter scale the models exhibited foundational failures in visual grounding, inaccurately extracting data points from the chart and demonstrating a severe logical disconnect by selecting a final option that contradicted its own flawed reasoning trace. Scaling to the 7-12B parameter range yielded significant improvements in visual parsing and logical architecture, enabling the model to formulate a valid and efficient problem-solving strategy; however, its execution remained brittle, as evidenced by minor arithmetic inaccuracies during the calculation phase. 

In contrast, the 27-32B model demonstrated highly robust and flawless execution across the entire visual-reasoning pipeline. It accurately extracted all relevant data, executed a precise multi-step calculation with the correct unit conversions, and logically arrived at the correct answer, ultimately highlighting that reliable, end-to-end performance on complex chart interpretation and mathematical reasoning tasks currently necessitates larger parameter scales.

\subsection{Error Analysis}

Analysis of the failure cases reveals clear patterns across model families and scales. We show an example using multi-lingual responses in Table \ref{tab:error_multilingual}. Smaller models (3-4B) frequently show weak visual grounding, evidenced by inaccurate extraction of data points, misunderstanding of core task objectives, and severe logical disconnects between intermediate reasoning traces and final answer selections. As capacity increases (7-12B), these foundational parsing issues largely diminish, with mid-sized models exhibiting improved vision-language alignment and the ability to formulate valid, multi-step problem-solving strategies. However, performance bottlenecks persist at this intermediate scale, predominantly manifesting as brittle mathematical execution where models falter on minor arithmetic inaccuracies despite sound logical frameworks. 

At the larger 27-32B scale, models successfully bridge these gaps, demonstrating robust data extraction and flawless end-to-end mathematical execution. Nonetheless, the persistent arithmetic fragility of intermediate models indicates that while structural visual understanding emerges at lower parameter counts, reliable execution of coupled visual-mathematical tasks remains highly dependent on increased model scale.

\section{Conclusion and future work}
This paper introduces \textit{FinVQA}, a multimodal Financial VQA benchmark covering English and five Indic languages such as Hindi, Bengali, Marathi, Gujarati, and Punjabi. It evaluates multilingual financial numerical problems along with their corresponding reasoning across 14 domains, structured into three difficulty levels (easy, moderate, hard) with diverse task formats.
 By integrating supervised fine-tuning with constraint-aware decoding,\textit{ FIND} enables rigorous and reliable assessment of numerical faithfulness, multimodal alignment, and structured decision-making in high-stakes financial settings.

As a future direction, we aim to investigate reward-based methods in reinforcement learning environments and undertake a more comprehensive analysis of multilingual financial reasoning, with a particular focus on examining how performance varies across languages and identifying the factors that influence reasoning quality in each linguistic setting.
\section{Limitations}
Despite demonstrating encouraging performance, the models exhibit several notable limitations. We observe frequent failures in verbal output quality, including repetitive generations, overly verbose responses, and inconsistent or unstable reasoning. In multiple cases, the reasoning presented by the model does not align with the final answer selected, indicating a disconnect between intermediate inference and answer selection. Additionally, inconsistent adherence to the required output format often results in multiple or ambiguous answer predictions, further reducing reliability. We employ two complementary techniques: constrained decoding and supervised fine-tuning which mitigate a subset of these issues.  However, several challenges remain unresolved:
\begin{itemize}
    \item \textbf{Multi-step reasoning consistency:} Difficulty in maintaining consistent and correct reasoning across multiple inference steps.
    \item \textbf{Visual information extraction and utilization:} Inability to accurately extract and effectively use information from the visual modality. 
    \item \textbf{Financial domain understanding:} Limited robustness in financial knowledge and conceptual understanding, especially in complex or multimodal scenarios.
    \item \textbf{Sample Coverage:} The current benchmark includes 18,900 multilingual instances generated from approximately 3,150 unique source questions translated across multiple languages. This construction supports systematic cross-lingual comparison by preserving semantic alignment across languages; however, it also implies that the diversity of underlying source instances is narrower than the total sample count alone might indicate. We view this as a practical first step toward multilingual financial benchmarking, and in future work we plan to expand the resource with more natively authored and semantically distinct questions to strengthen linguistic diversity, cultural nuance, and benchmark coverage.
    \item \textbf{Closed-Source Baselines}: Our evaluation does not yet include a wider set of closed-source baselines or more extensive Chain-of-Thought analysis, mainly due to limited computational and API resources. As the study was conducted without dedicated external funding, large-scale experimentation with high-cost proprietary models was not feasible. We will clarify this constraint in the camera-ready version and discuss the resulting evaluation gap accordingly. 
\end{itemize}

\section{Ethical Considerations}
\textit{FIND} targets financial numerical and multimodal reasoning, a high-stakes domain where incorrect or misleading outputs can have real economic consequences. A primary data-level ethical concern is the risk of misuse, as models evaluated on \textit{FIND} may be perceived as reliable financial advisors despite the benchmark not being designed for prescriptive or personalized financial decision-making. Accordingly, \textit{FIND} is explicitly intended for evaluation and research only and contains no real user data or individualized financial scenarios.

From a linguistic perspective, although \textit{FIND} covers multiple Indic languages, uneven model performance across languages may reinforce existing disparities in financial literacy if not carefully interpreted. To mitigate this risk, the dataset maintains balanced difficulty levels and domain coverage across languages, and emphasizes language-wise reporting rather than aggregate claims.

At the model level, \textit{FIND} exposes a critical ethical issue: fluency without correctness. Large reasoning models often generate confident explanations even when numerical reasoning or formula application is incorrect, posing a risk of user over-trust in financial contexts. Additionally, vision–language models frequently under-utilize visual financial evidence, relying instead on textual priors, which can lead to systematically flawed multimodal reasoning.

Finally, while constraint-aware decoding improves format compliance, it may also mask underlying reasoning errors, creating an illusion of reliability. Overall, \textit{FIND }highlights that model scale and structured outputs alone do not guarantee trustworthy financial reasoning, underscoring the necessity of human oversight, transparent evaluation, and cautious interpretation in any downstream use.

\section{Acknowledgement}
All authors sincerely acknowledge the invaluable contributions of the dataset annotators Paavne, Jheel, Yajant, and Alekhya who were instrumental in the successful completion of the annotation process through their dedicated, proactive, and sustained efforts. 

\bibliography{custom}

\appendix
\section{Dataset Details}
\label{sec:data}
\subsection{Domain Coverage and Definition}

The dataset is systematically organized across \textbf{14 financial, economic, and decision-centric domains}, each targeting distinct reasoning competencies relevant to real-world financial understanding. Below, we define each domain along with representative attributes illustrating its scope.

\begin{itemize}

\item \textbf{Advance Accounting (2,856 samples):}  
Focuses on higher-order accounting treatments requiring multi-step reasoning.  
\textit{Attributes:} partnership admission and retirement, goodwill valuation, amalgamation, revaluation accounts, company final accounts.

\item \textbf{Core Accountancy (1,290 samples):}  
Covers procedural accounting fundamentals emphasizing accuracy and rule-based reasoning.  
\textit{Attributes:} journal entries, ledger posting, trial balance preparation, error rectification.

\item \textbf{Fundamental Accountancy (1,356 samples):}  
Introduces basic accounting principles and conceptual foundations.  
\textit{Attributes:} matching concept, accrual accounting, depreciation, asset–liability classification.

\item \textbf{Finance on Accountancy (2,274 samples):}  
Integrates financial analysis with accounting data for decision-making.  
\textit{Attributes:} ratio analysis, capital structure decisions, break-even analysis, financial statement interpretation.

\item \textbf{Financial Mathematics (1,236 samples):}  
Targets numerical and quantitative reasoning in financial contexts.  
\textit{Attributes:} simple and compound interest, annuities, time value of money, percentage change, ratio computation.

\item \textbf{Basic Statistics (1,374 samples):}  
Covers descriptive and introductory inferential statistics applied to financial data.  
\textit{Attributes:} mean, median, mode, probability basics, data interpretation from tables and charts.

\item \textbf{Business Economics (270 samples):}  
Examines firm-level economic reasoning and managerial decision contexts.  
\textit{Attributes:} cost functions, revenue analysis, production decisions, profit maximization.

\item \textbf{Fundamental Economics (978 samples):}  
Introduces core microeconomic concepts and market mechanisms.  
\textit{Attributes:} demand and supply analysis, elasticity, consumer behavior, equilibrium pricing.

\item \textbf{Price and Discrimination (300 samples):}  
Focuses on pricing strategies and market regulation.  
\textit{Attributes:} price floors and ceilings, degrees of price discrimination, monopolistic pricing.

\item \textbf{Means of Production (804 samples):}  
Addresses the role of production factors in economic output.  
\textit{Attributes:} land, labor, capital, entrepreneurship, factor remuneration.

\item \textbf{Decision Making (2,784 samples):}  
Captures applied reasoning under uncertainty and constraints.  
\textit{Attributes:} logical reasoning, risk assessment, optimization, scenario-based choices.

\item \textbf{Taxation (1,962 samples):}  
Covers compliance-oriented financial reasoning and tax computation.  
\textit{Attributes:} income tax calculation, GST, exemptions, deductions, tax slabs.

\item \textbf{B.Com (1,266 samples):}  
Represents integrated undergraduate-level commerce problems combining multiple disciplines.  
\textit{Attributes:} mixed accounting–finance problems, numerical aptitude, applied economics.

\item \textbf{Dimensions of VUCAFU (150 samples):}  
Targets modern strategic reasoning under dynamic economic conditions.  
\textit{Attributes:} volatility, uncertainty, complexity, ambiguity, fragility, unpredictability.

\end{itemize}

\subsection{Difficulty Stratification}

Each domain is uniformly stratified across three levels of cognitive complexity:
\textbf{Easy (6,282 samples)} targeting conceptual recall and single-step reasoning,
\textbf{Moderate (6,384 samples)} emphasizing applied multi-step computation,
and \textbf{Hard (6,234 samples)} requiring complex, multi-hop financial reasoning and decision analysis. 

\section{Text--Image Annotation Guidelines}
\label{label:dataset_annotation}
\subsection{Textual Annotation Principles}\label{guidelines}
To ensure semantic accuracy, consistency, and reasoning fidelity in textual annotations, annotators must adhere to the following principles:

\begin{enumerate}
    \item \textbf{Semantic Faithfulness:}  
    Text annotations must accurately reflect the intended meaning of the input question or content, without introducing unsupported assumptions or hallucinated information.

    \item \textbf{Conceptual Correctness:}  
    Ensure that all domain-specific concepts (e.g., financial, economic, statistical) are used correctly and consistently, adhering to standard definitions and conventions.

    \item \textbf{Clarity and Precision:}  
    Textual explanations should be concise, unambiguous, and logically structured, avoiding unnecessary verbosity while preserving essential reasoning steps.

    \item \textbf{Reasoning Transparency:}  
    When explanations are required, intermediate reasoning must align coherently with the final answer, ensuring traceability between assumptions, computations, and conclusions.

    \item \textbf{Context Appropriateness:}  
    Incorporate only the contextual information necessary to interpret or solve the task. Irrelevant or speculative details should be excluded.

    \item \textbf{Consistency and Formatting:}  
    Maintain a uniform annotation style across all samples, including consistent terminology, notation, and response structure.

    \item \textbf{Bias and Neutrality Check:}  
    Avoid subjective, cultural, or demographic bias in textual descriptions and reasoning. All annotations should remain neutral and task-focused.

\end{enumerate}

\subsection{Image Annotation and Validation Rules}
The following rules govern the annotation and validation of visual inputs to ensure alignment, quality, and reliability in multimodal settings:

\begin{enumerate}
    \item \textbf{Rule 1: Text--Image Semantic Alignment} \\
    The image must provide visual evidence that is directly relevant to the associated text or question. Visual content should support, not contradict, the intended reasoning task.

    \item \textbf{Rule 2: Visual Evidence Sufficiency} \\
    Annotate only those images where the necessary information is visually discernible. Images that are ambiguous, misleading, or insufficient to support reasoning should be excluded.

    \item \textbf{Rule 3: No Embedded or Overlay Text} \\
    Images must not contain embedded text, labels, watermarks, or annotations, as these can introduce unintended shortcuts or textual bias.

    \item \textbf{Rule 4: Visual Quality and Readability} \\
    Images should be clear, properly cropped, and well-illuminated. Low-resolution, blurred, or cluttered visuals that hinder interpretation are not permitted.

    \item \textbf{Rule 5: Numerical and Structural Integrity} \\
    For charts, tables, or financial visuals, ensure that axes, symbols, and numerical values are visually intact and correctly rendered, without distortion or truncation.

    \item \textbf{Rule 6: Human and Object Realism} \\
    Any depicted humans or objects must appear natural and undistorted. Images exhibiting anatomical inconsistencies or AI-induced artifacts should be rejected.

\end{enumerate}

\vspace{1em}

These guidelines ensure that textual and visual annotations remain semantically aligned, logically coherent, and visually reliable, thereby enabling robust multimodal reasoning, evaluation, and model generalisation.

\begin{table*}[t]
\centering
\caption{Multilingual Sample Instance (Sample ID: 3) Across Six Languages}
\scriptsize
\begin{tabular}{|l|p{0.85\textwidth}|}
\hline
\textbf{Language} & \textbf{Details} \\
\hline

\textbf{English} &
\parbox[t]{\linewidth}{
\textbf{Sub-Domain:} Core Accountancy\\
\textbf{Question:} If average inventory is ₹ 1,25,000 and closing inventory is ₹ 10,000 less than opening inventory, then the value of closing inventory will be.\\
\textbf{Options:} [A] ₹ 1,35,000 \quad [B] ₹ 1,15,000 \quad [C] ₹ 1,30,000 \quad [D] ₹ 1,20,000\\
\textbf{Answer:} Option [D]\\
\textbf{Reasoning:} Average Inventory = (Opening + Closing)/2. Let Opening = X, Closing = X − 10,000.\\
1,25,000 = (2X − 10,000)/2 $\Rightarrow$ 2,50,000 = 2X − 10,000 $\Rightarrow$ 2,60,000 = 2X $\Rightarrow$ X = 1,30,000.\\
Closing = 1,20,000.
}\\
\hline

\textbf{Hindi} &
\parbox[t]{\linewidth}{
\textbf{Sub-Domain:} {\hindifont कोर अकाउंटेंसी}\\
\textbf{Question:} {\hindifont यदि औसत इन्वेंटरी ₹१,२५,००० है और समापन इन्वेंटरी, उद्घाटन इन्वेंटरी से ₹१०,००० कम है, तो समापन इन्वेंटरी का मूल्य क्या होगा?}\\
\textbf{Options:} {\hindifont [क] ₹१,३५,००० \quad [ख] ₹१,१५,००० \quad [ग] ₹१,३०,००० \quad [घ] ₹१,२०,०००}\\
\textbf{Answer:} {\hindifont विकल्प [घ]}\\
\textbf{Reasoning:} {\hindifont औसत इन्वेंटरी = (उद्घाटन इन्वेंटरी + समापन इन्वेंटरी)/२। मान लें उद्घाटन इन्वेंटरी = \textenglish{X}, तब समापन इन्वेंटरी = \textenglish{X} − ₹१०,०००।}\\
{\hindifont ₹१,२५,००० = (२\textenglish{X} − ₹१०,०००)/२ $\Rightarrow$ ₹२,५०,००० = २\textenglish{X} − ₹१०,००० $\Rightarrow$ ₹२,६०,००० = २\textenglish{X} $\Rightarrow$ \textenglish{X} = ₹१,३०,०००।}\\
{\hindifont समापन इन्वेंटरी = ₹१,२०,०००।}
}\\
\hline

\textbf{Bengali} &
\parbox[t]{\linewidth}{
\textbf{Sub-Domain:} {\bengalifont কোর অ্যাকাউন্টেন্সি}\\
\textbf{Question:} {\bengalifont যদি গড় ইনভেন্টরি ₹১,২৫,০০০ হয় এবং সমাপনী ইনভেন্টরি উদ্বোধনী ইনভেন্টরির থেকে ₹১০,০০০ কম হয়, তবে সমাপনী ইনভেন্টরির মূল্য কত হবে?}\\
\textbf{Options:} {\bengalifont [ক] ₹১,৩৫,০০০ \quad [খ] ₹১,১৫,০০০ \quad [গ] ₹১,৩০,০০০ \quad [ঘ] ₹১,২০,০০০}\\
\textbf{Answer:} {\bengalifont বিকল্প [ঘ]}\\
\textbf{Reasoning:} {\bengalifont গড় ইনভেন্টরি = (উদ্বোধনী ইনভেন্টরি + সমাপনী ইনভেন্টরি)/২। ধরা যাক উদ্বোধনী ইনভেন্টরি = \textenglish{X}, তবে সমাপনী ইনভেন্টরি = \textenglish{X} − ₹১০,০০০।}\\
{\bengalifont ₹১,২৫,০০০ = (২\textenglish{X} − ₹১০,০০০)/২ $\Rightarrow$ ₹২,৫০,০০০ = ২\textenglish{X} − ₹১০,০০০ $\Rightarrow$ ₹২,৬০,০০০ = ২\textenglish{X} $\Rightarrow$ \textenglish{X} = ₹১,৩০,০০০।}\\
{\bengalifont সমাপনী ইনভেন্টরি = ₹১,২০,০০০।}
}\\
\hline

\textbf{Marathi} &
\parbox[t]{\linewidth}{
\textbf{Sub-Domain:} {\marathifont कोअर अकाउंटन्सी}\\
\textbf{Question:} {\marathifont जर सरासरी साठा ₹१,२५,००० असेल आणि समापन साठा उद्घाटन साठ्यापेक्षा ₹१०,००० ने कमी असेल, तर समापन साठ्याची किंमत किती असेल?}\\
\textbf{Options:} {\marathifont [क] ₹१,३५,००० \quad [ख] ₹१,१५,००० \quad [ग] ₹१,३०,००० \quad [घ] ₹१,२०,०००}\\
\textbf{Answer:} {\marathifont पर्याय [घ]}\\
\textbf{Reasoning:} {\marathifont सरासरी साठा = (उद्घाटन साठा + समापन साठा)/२। गृहित धरा उद्घाटन साठा = \textenglish{X}, तर समापन साठा = \textenglish{X} − ₹१०,०००।}\\
{\marathifont ₹१,२५,००० = (२\textenglish{X} − ₹१०,०००)/२ $\Rightarrow$ ₹२,५०,००० = २\textenglish{X} − ₹१०,००० $\Rightarrow$ ₹२,६०,००० = २\textenglish{X} $\Rightarrow$ \textenglish{X} = ₹१,३०,०००।}\\
{\marathifont समापन साठा = ₹१,२०,०००।}
}\\
\hline

\textbf{Gujarati} &
\parbox[t]{\linewidth}{
\textbf{Sub-Domain:} {\gujaratifont કોર એકાઉન્ટન્સી}\\
\textbf{Question:} {\gujaratifont જો સરેરાશ ઇન્વેન્ટરી ₹૧,૨૫,૦૦૦ છે અને ક્લોઝિંગ ઇન્વેન્ટરી ઓપનિંગ ઇન્વેન્ટરી કરતાં ₹૧૦,૦૦૦ ઓછી છે, તો ક્લોઝિંગ ઇન્વેન્ટરીનું મૂલ્ય કેટલું હશે?}\\
\textbf{Options:} {\gujaratifont [ક] ₹૧,૩૫,૦૦૦ \quad [ખ] ₹૧,૧૫,૦૦૦ \quad [ગ] ₹૧,૩૦,૦૦૦ \quad [ઘ] ₹૧,૨૦,૦૦૦}\\
\textbf{Answer:} {\gujaratifont વિકલ્પ [ઘ]}\\
\textbf{Reasoning:} {\gujaratifont સરેરાશ ઇન્વેન્ટરી = (ઓપનિંગ ઇન્વેન્ટરી + ક્લોઝિંગ ઇન્વેન્ટરી)/૨। માનીએ કે ઓપનિંગ ઇન્વેન્ટરી = \textenglish{X}, તો ક્લોઝિંગ ઇન્વેન્ટરી = \textenglish{X} − ₹૧૦,૦૦૦।}\\
{\gujaratifont ₹૧,૨૫,૦૦૦ = (૨\textenglish{X} − ₹૧૦,૦૦૦)/૨ $\Rightarrow$ ₹૨,૫૦,૦૦૦ = ૨\textenglish{X} − ₹૧૦,૦૦૦ $\Rightarrow$ ₹૨,૬૦,૦૦૦ = ૨\textenglish{X} $\Rightarrow$ \textenglish{X} = ₹૧,૩૦,૦૦૦।}\\
{\gujaratifont ક્લોઝિંગ ઇન્વેન્ટરી = ₹૧,૨૦,૦૦૦।}
}\\
\hline

\textbf{Tamil} &
\parbox[t]{\linewidth}{
\textbf{Sub-Domain:} {\tamilfont கோர் அக்கௌண்டன்சி}\\
\textbf{Question:} {\tamilfont சராசரி சரக்கு ₹௧,௨௫,௦௦௦ ஆகவும், நிறைவுச் சரக்கு தொடக்கச் சரக்கை விட ₹௧௦,௦௦௦ குறைவாக இருந்தால், நிறைவுச் சரக்கின் மதிப்பு எவ்வளவு?}\\
\textbf{Options:} {\tamilfont [௧] ₹௧,௩௫,௦௦௦ \quad [௨] ₹௧,௧௫,௦௦௦ \quad [௩] ₹௧,௩௦,௦௦௦ \quad [௪] ₹௧,௨௦,௦௦௦}\\
\textbf{Answer:} {\tamilfont விருப்பம் [௪]}\\
\textbf{Reasoning:} {\tamilfont சராசரி சரக்கு = (தொடக்கச் சரக்கு + நிறைவுச் சரக்கு)/௨. தொடக்கச் சரக்கு = \textenglish{X} எனக் கொள்ளுங்கள். அப்போது நிறைவுச் சரக்கு = \textenglish{X} − ₹௧௦,௦௦௦.}\\
{\tamilfont ₹௧,௨௫,௦௦௦ = (௨\textenglish{X} − ₹௧௦,௦௦௦)/௨ $\Rightarrow$ ₹௨,௫௦,௦௦௦ = ௨\textenglish{X} − ₹௧௦,௦௦௦ $\Rightarrow$ ₹௨,௬௦,௦௦௦ = ௨\textenglish{X} $\Rightarrow$ \textenglish{X} = ₹௧,௩௦,௦௦௦.}\\
{\tamilfont நிறைவுச் சரக்கு = ₹௧,௨௦,௦௦௦.}
}\\
\hline
\end{tabular}
\label{tab:multilingual_sample_inventory}
\end{table*}
\begin{figure*}[!h]
    \centering
    \includegraphics[width=\linewidth]{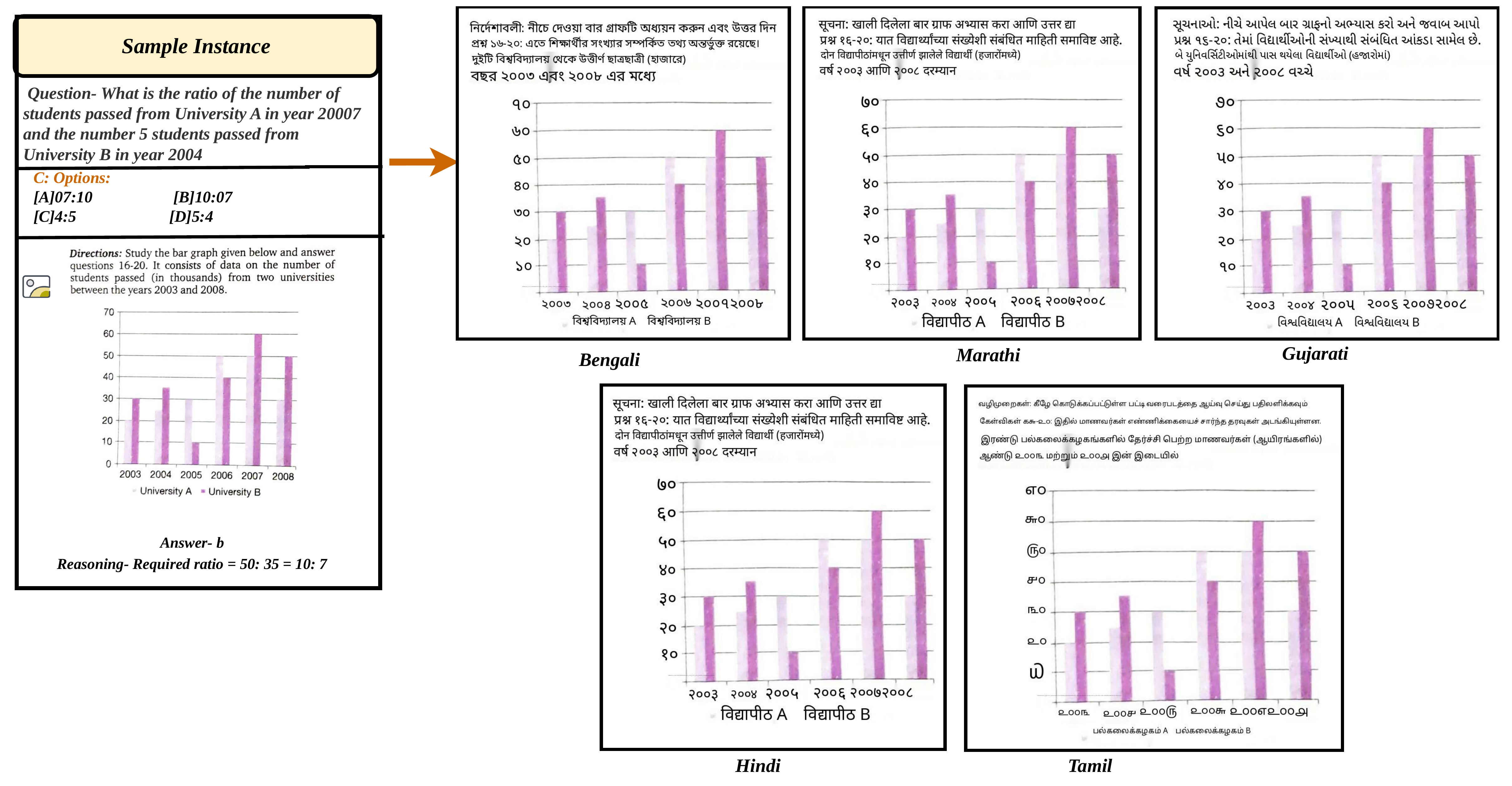}
    \caption{OCR-Based Text Conversion.}
    \label{fig:ocr_based_text}
\end{figure*}

\section{Prompts}\label{label:prompts}

\begin{promptbox}{Prompt used for Zero-shot inference variation}
\textbf{System Role:} You are a helpful assistant. \\
\textbf{Instruction:} You will be given a multiple-choice question (MCQ). Solve the question and choose the correct option from the given choices (A, B, C, or D). Explain your reasoning step by step before providing your final answer.

Format your output as:\\
<reasoning> reasoning\_text </reasoning>\\
<answer> Option [A] or [B] or [C] or [D] </answer>

The question and the options are given in \{langauge\} language. Provide the answer and reasoning in the same language - \{langauge\}, and the tags in English as shown above.
The question and the possible answers are as follows:\\
Question: \{question\}
Options: \{options\}
\end{promptbox}

\begin{promptbox}{Prompt used for Constrained decoding variation}
\textbf{System Role:} You are a helpful assistant. \\
\textbf{Instruction:} You will be given a multiple-choice question (MCQ). Solve the question and choose the correct option from the given choices (A, B, C, or D). Explain your reasoning step by step before providing your final answer.

The output should be a single letter only corresponding to the correct option: A or B or C or D\\
Do not output anything else.\\
The question and the options are given in \{langauge\} language. Provide the answer option in the same language - \{langauge\}. The question and the possible answers are as follows:\\
Question: \{question\}
Options: \{options\}
\end{promptbox}

\section{Human Evaluation Details}\label{human_eval_guidelines}
Human evaluation was conducted exclusively under the zero-shot setting (Variation-1). This restriction is intentional: in the zero-shot regime, models generate complete outputs comprising reasoning traces, option selection, and final answer justification, enabling a faithful assessment of intermediate reasoning quality. In contrast, under supervised fine-tuning and constraint decoding settings, the generation process becomes restricted, with models predominantly producing option letters without explicit reasoning traces. This limits the ability to evaluate logical coherence, and step-wise correctness. Consequently, the proposed human evaluation framework is applied solely to zero-shot outputs.
To ensure balanced multilingual evaluation, we sample 100 instances per language, and all scores reported in Table~\ref{tab:financial_reasoning_human_eval} are derived from this controlled human annotation protocol. The evaluation follows a structured, rubric-driven methodology designed to capture multiple dimensions of financial reasoning beyond surface-level correctness. Specifically, the rubric evaluates six core criteria:
\begin{itemize}
    \item Financial Domain Understanding, assessing whether the model correctly captures domain-specific financial concepts.
    \item Problem Interpretation and Assumption Validity, measuring the correctness of question comprehension and the plausibility of implicit assumptions.
    \item Mathematical Correctness, verifying numerical accuracy and computational validity. 
    \item Formula Selection and Application, evaluating whether appropriate financial or mathematical formulations are correctly identified and applied.
    \item Reasoning Consistency and Answer Selection, assessing internal logical coherence and alignment between reasoning and final answer.
    \item Formatting and Output Compliance, ensuring adherence to the prescribed response structure.
\end{itemize}

This evaluation strategy enables a fine-grained, reasoning-centric analysis, moving beyond exact-match metrics to assess the faithfulness, interpretability, and practical usability of model outputs in financial QA. The results in Table \ref{tab:financial_reasoning_human_eval} reveal a consistent scaling trend across all model families, where performance improves with increasing parameter size. For instance, the average score for Qwen2.5VL increases from 0.46 (3B) to 0.70 (7B) and 0.94 (32B); Gemma 3 improves from 0.60 (4B) to 0.80 (12B) and 0.96 (27B); and Qwen3VL rises from 0.66 (4B) to 0.85 (8B) and 0.98 (32B). This progression indicates that larger models not only achieve higher answer accuracy but also exhibit substantially improved reasoning fidelity, formula grounding, and structural compliance.

A metric-wise analysis further highlights that Formatting and Output Compliance consistently achieves near-perfect scores for medium and large models, often reaching 1.00, indicating strong adherence to output constraints. Similarly, Formula Selection and Application and Reasoning Consistency demonstrate significant gains with scale, with several large models attaining perfect scores, reflecting robust capability in structured financial reasoning. In contrast, Financial Domain Understanding remains comparatively challenging for smaller models (e.g., 0.13 for Qwen2.5VL-3B, 0.24 for Gemma 3-4B, and 0.32 for Qwen 3VL-4B), suggesting that such models often lack deep semantic grounding despite occasionally producing correct outputs.

Overall, this human evaluation framework provides a rigorous, multi-dimensional assessment of financial QA systems, explicitly capturing reasoning quality, numerical faithfulness, and usability. By leveraging a multilingual, balanced sample (100 instances per language) and a carefully designed rubric, the analysis offers a comprehensive diagnostic perspective, complementing automatic metrics and enabling a more reliable evaluation of model behavior in high-stakes financial reasoning tasks.

\begin{figure*}[t]
\centering

\begin{minipage}{0.48\textwidth}
    \centering
    \textbf{Text-only}
\end{minipage}
\hfill
\begin{minipage}{0.48\textwidth}
    \centering
    \textbf{Text + Image}
\end{minipage}

\vspace{0.2cm}

\begin{subfigure}[t]{0.48\textwidth}
    \centering
    \includegraphics[width=\linewidth]{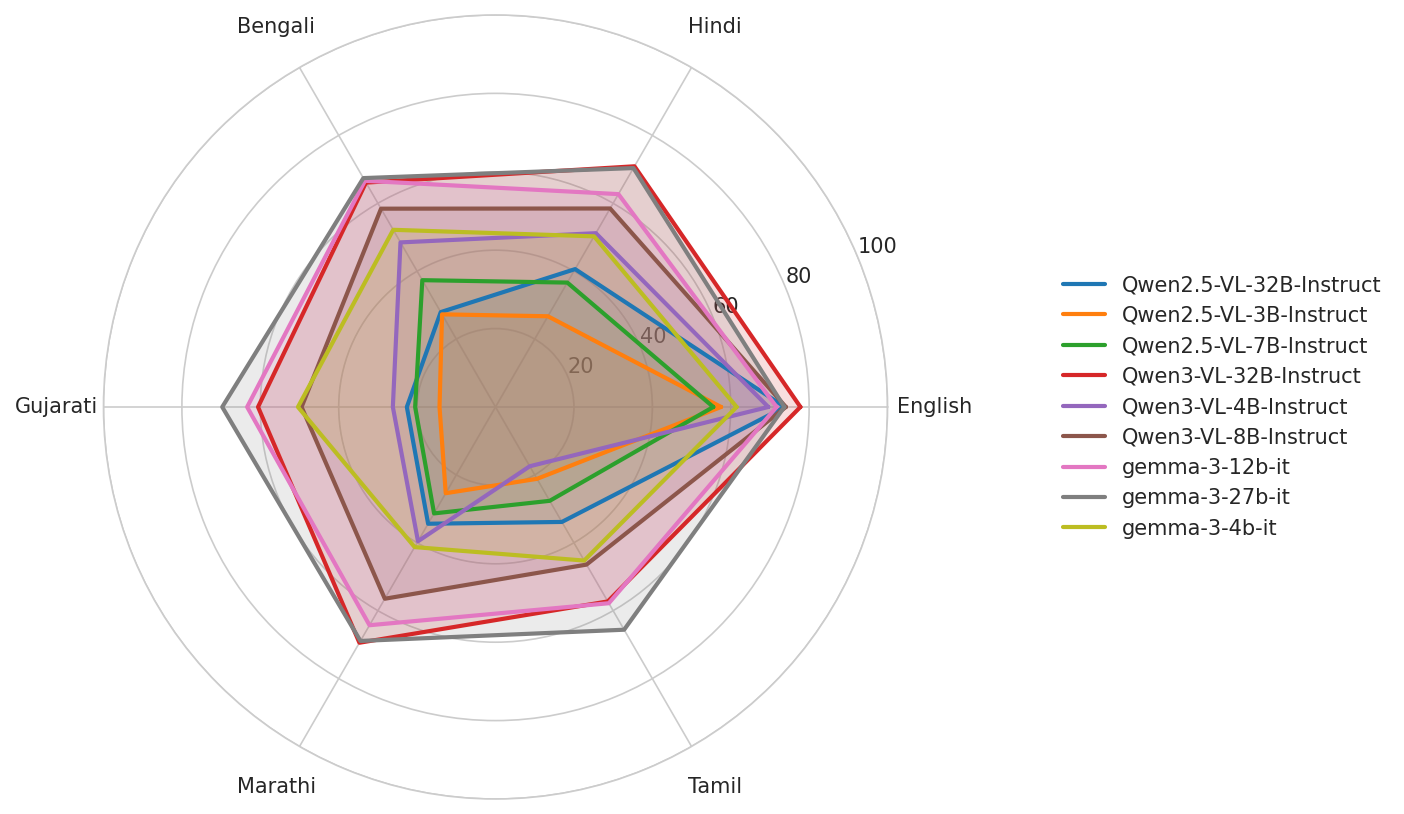}
    \caption{Zero-shot Inference}
\end{subfigure}
\hfill
\begin{subfigure}[t]{0.48\textwidth}
    \centering
    \includegraphics[width=\linewidth]{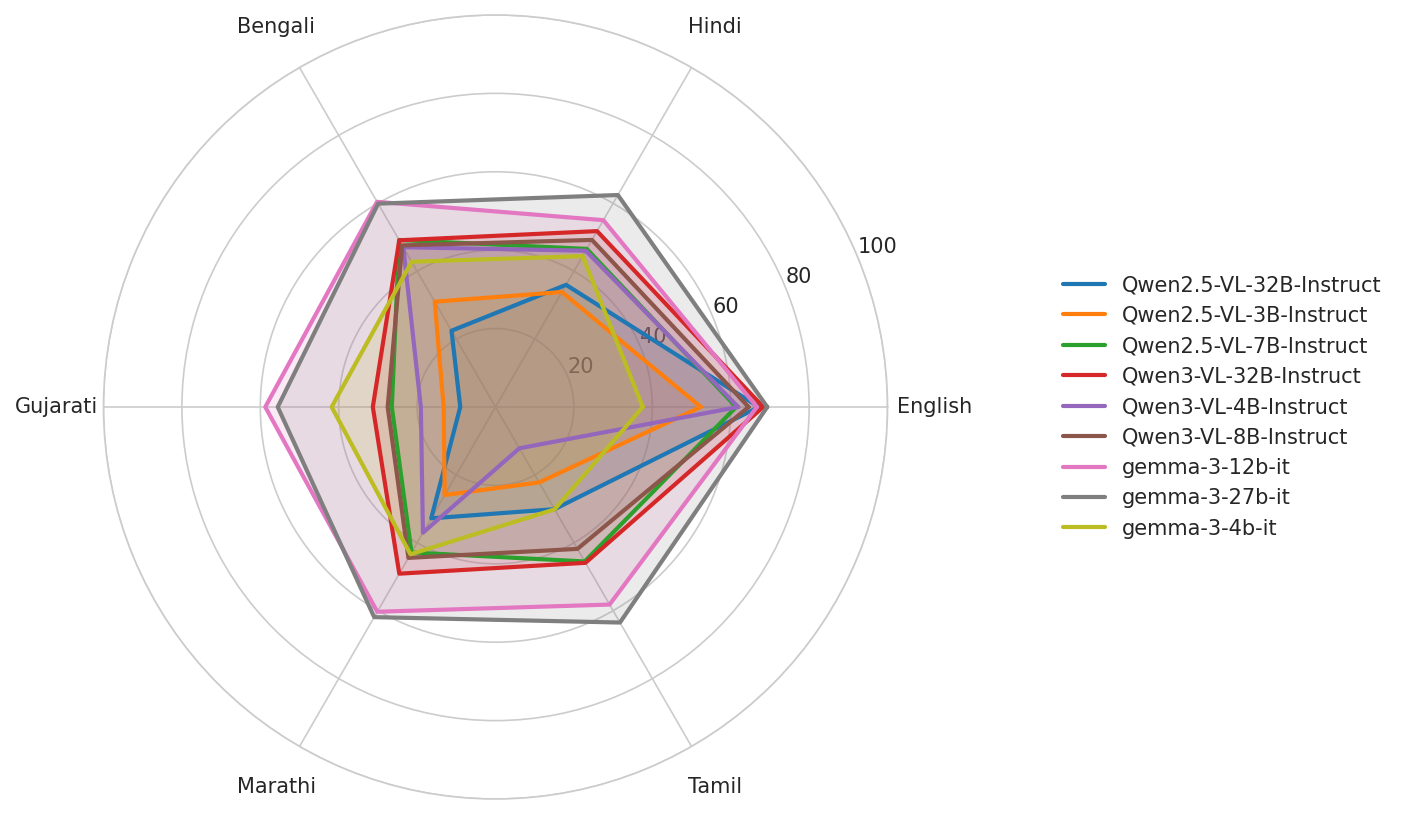}
    \caption{Zero-shot Inference}
\end{subfigure}

\vspace{0.25cm}

\begin{subfigure}[t]{0.48\textwidth}
    \centering
    \includegraphics[width=\linewidth]{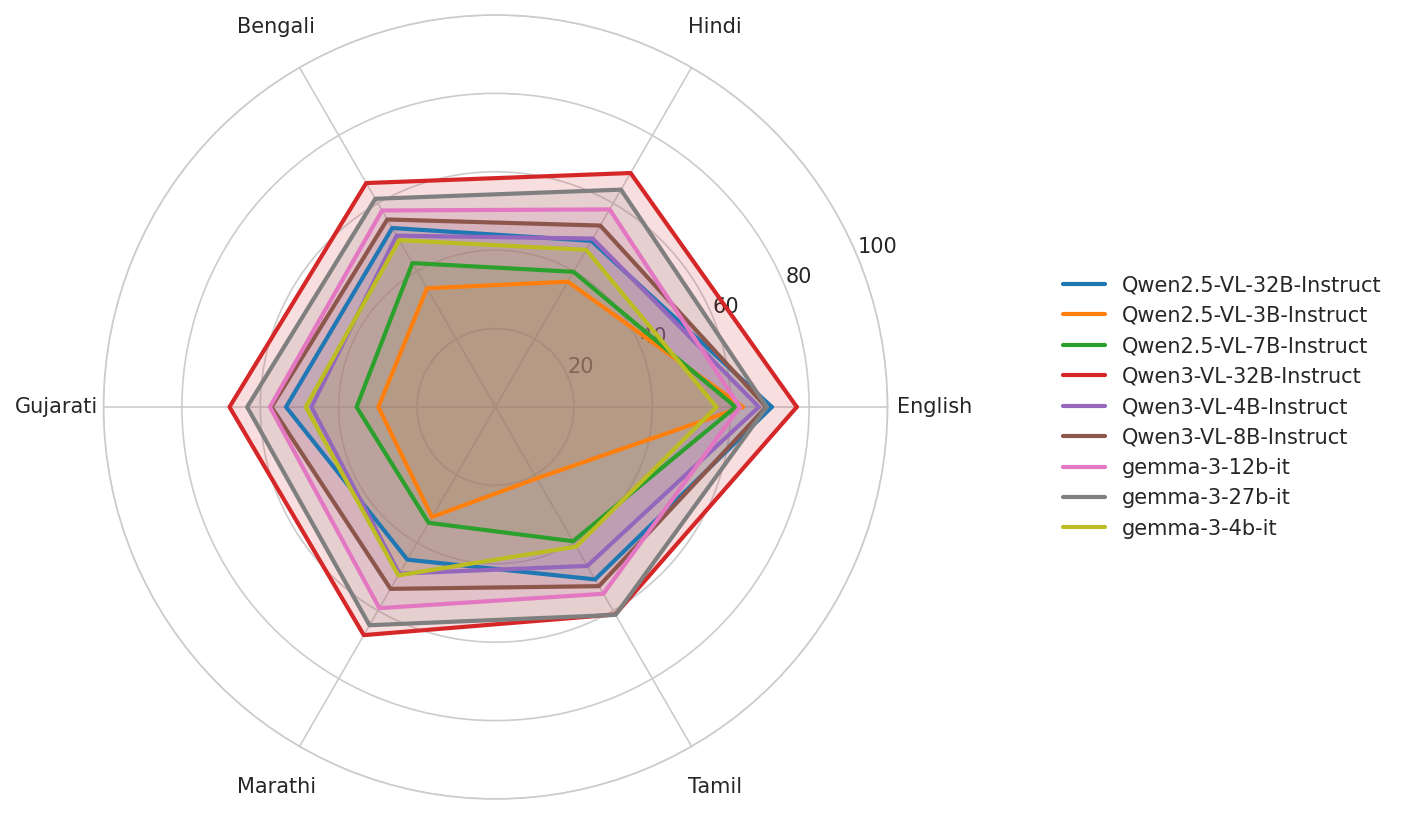}
    \caption{Constrained decoding}
\end{subfigure}
\hfill
\begin{subfigure}[t]{0.48\textwidth}
    \centering
    \includegraphics[width=\linewidth]{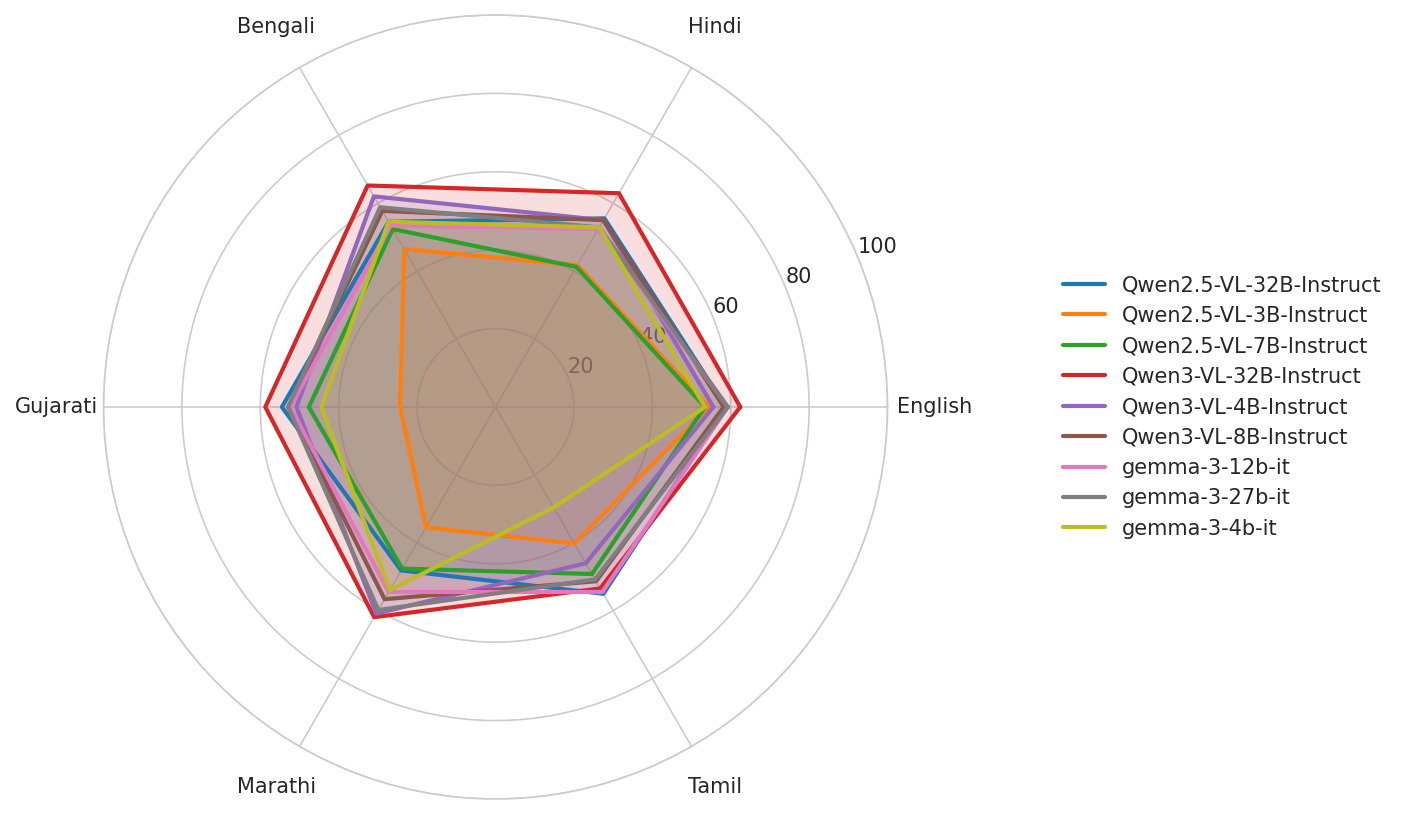}
    \caption{Constrained decoding}
\end{subfigure}

\vspace{0.25cm}

\begin{subfigure}[t]{0.48\textwidth}
    \centering
    \includegraphics[width=\linewidth]{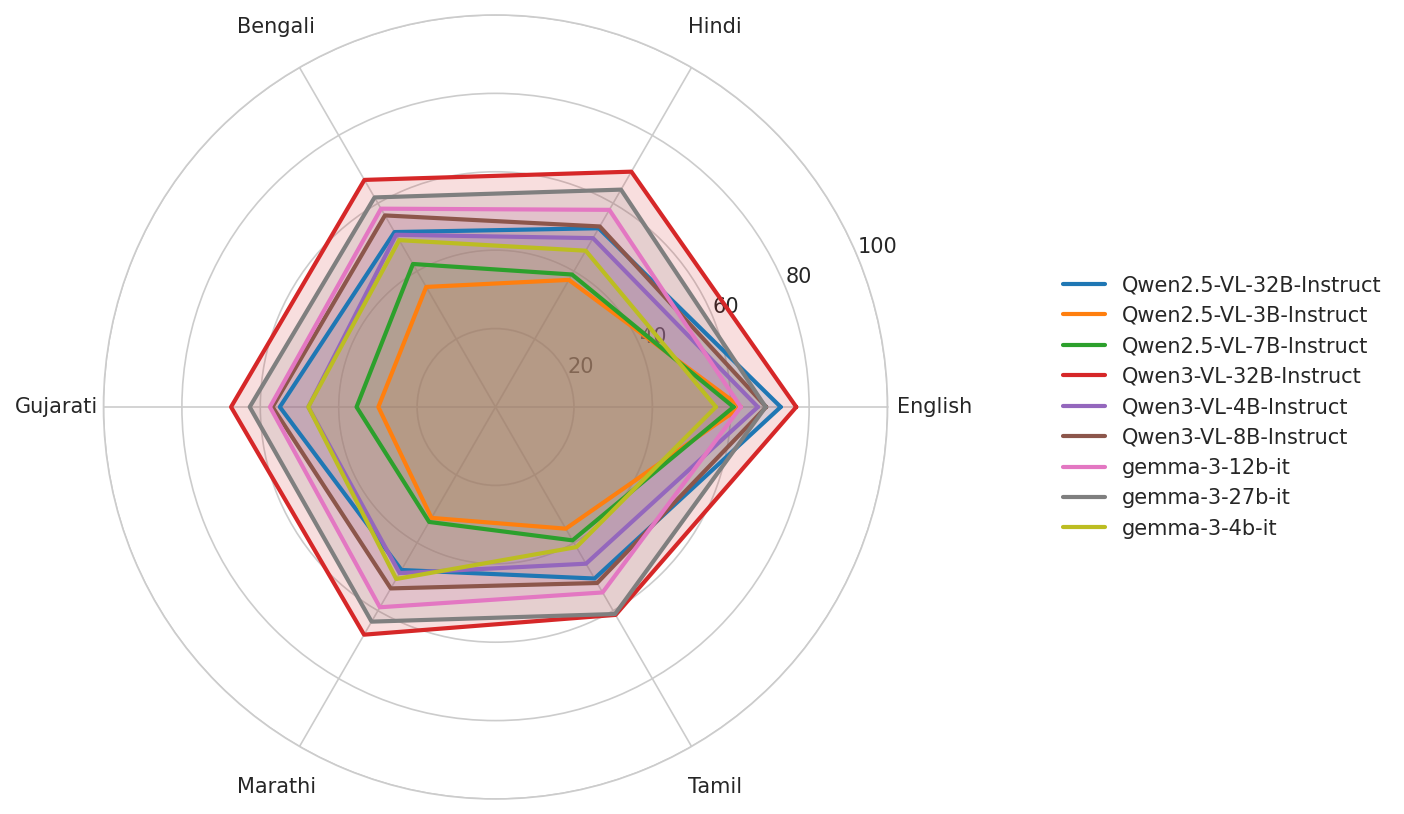}
    \caption{SFT + constrained decoding}
\end{subfigure}
\hfill
\begin{subfigure}[t]{0.48\textwidth}
    \centering
    \includegraphics[width=\linewidth]{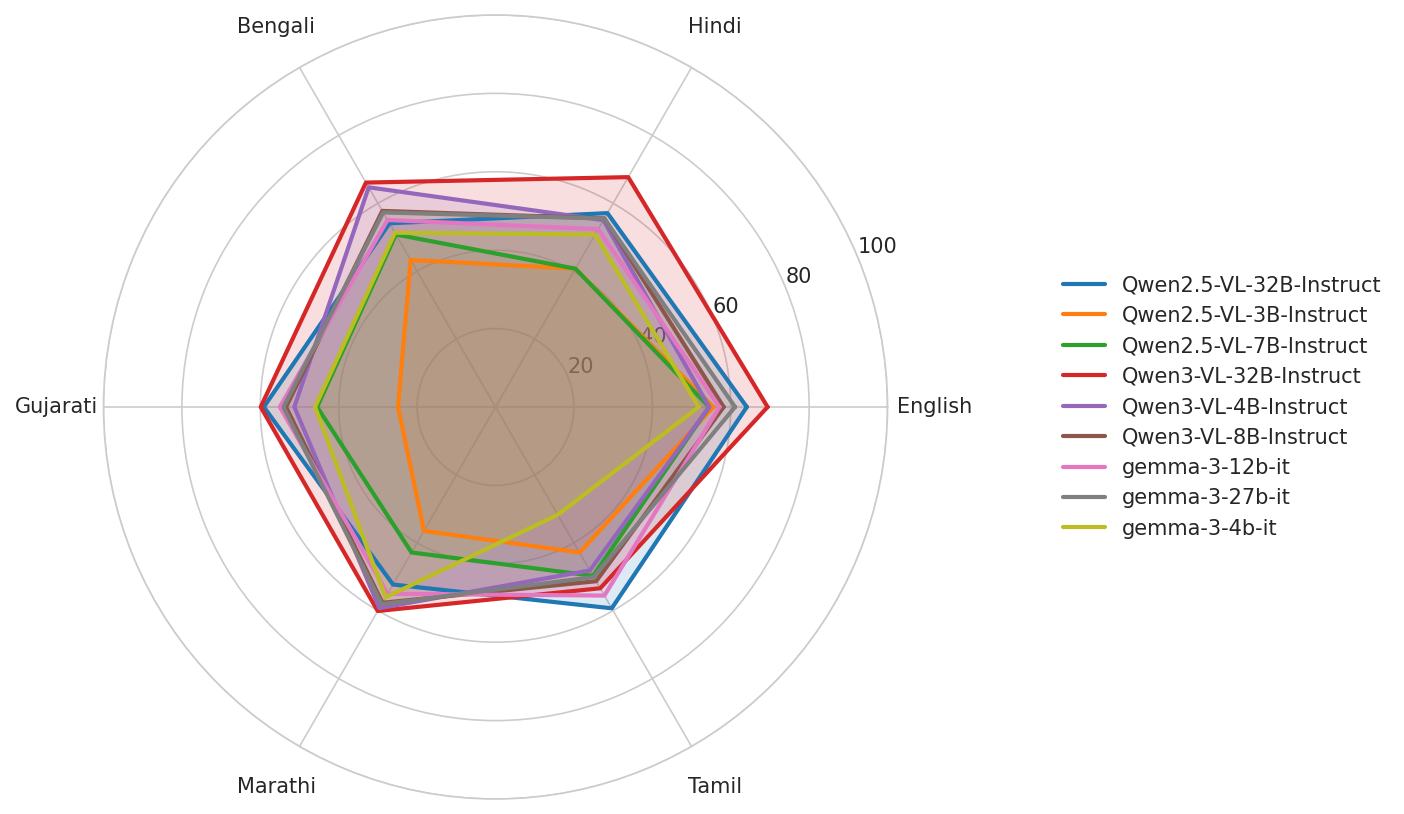}
    \caption{SFT + constrained decoding}
\end{subfigure}

\caption{
Comparison of model performance across inference and training strategies.
Rows correspond to different generation strategy,
while columns contrast text-only and text+image inputs.
}
\label{fig:experiment_results_fig_option_1}
\end{figure*}

\begin{figure*}[t]
\centering

\begin{minipage}{0.48\textwidth}
    \centering
    \textbf{Text-only}
\end{minipage}
\hfill
\begin{minipage}{0.48\textwidth}
    \centering
    \textbf{Text + Image}
\end{minipage}

\vspace{0.2cm}

\begin{subfigure}[t]{0.48\textwidth}
    \centering
    \includegraphics[width=\linewidth]{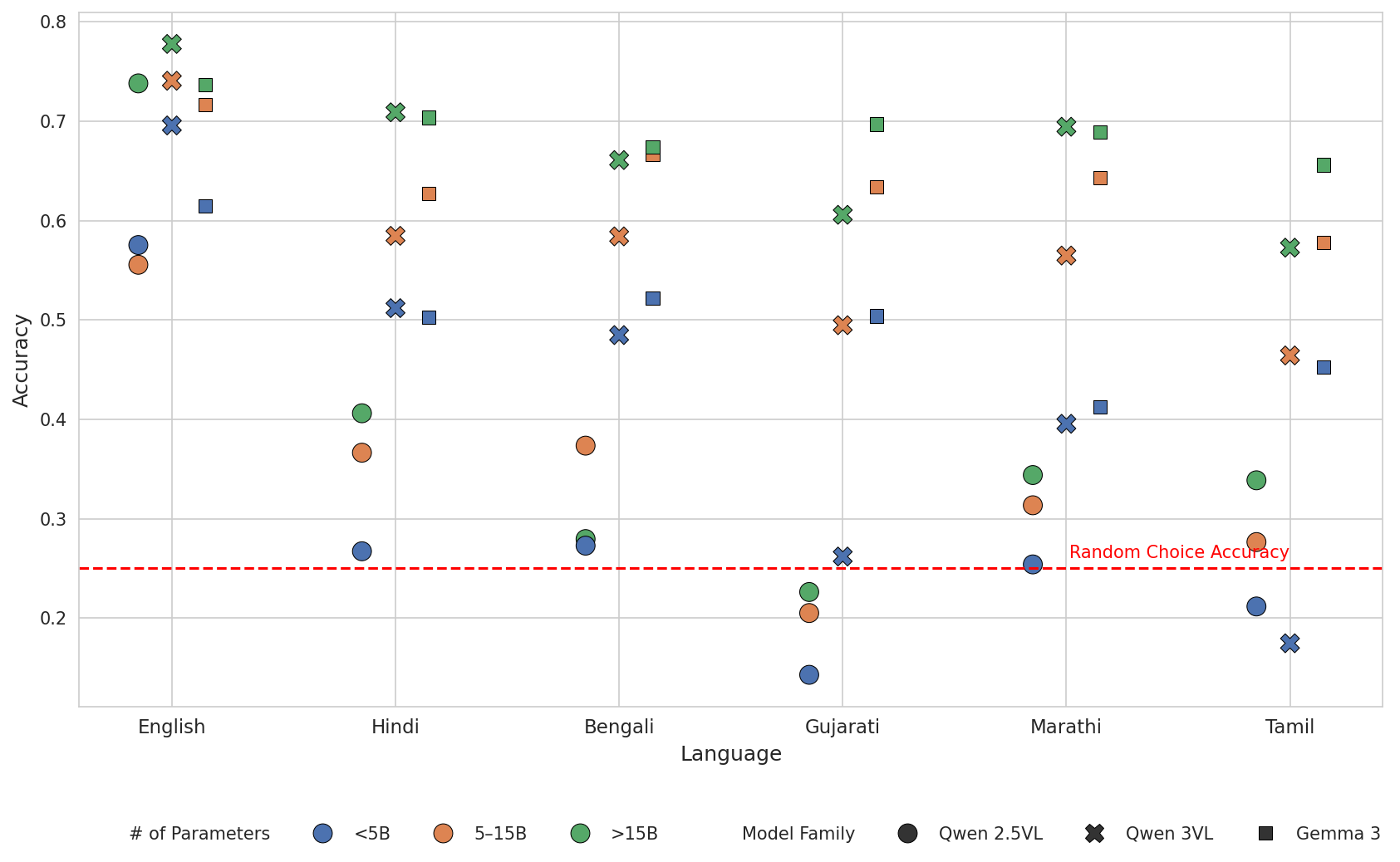}
    \caption{Zero-shot Inference}
\end{subfigure}
\hfill
\begin{subfigure}[t]{0.48\textwidth}
    \centering
    \includegraphics[width=\linewidth]{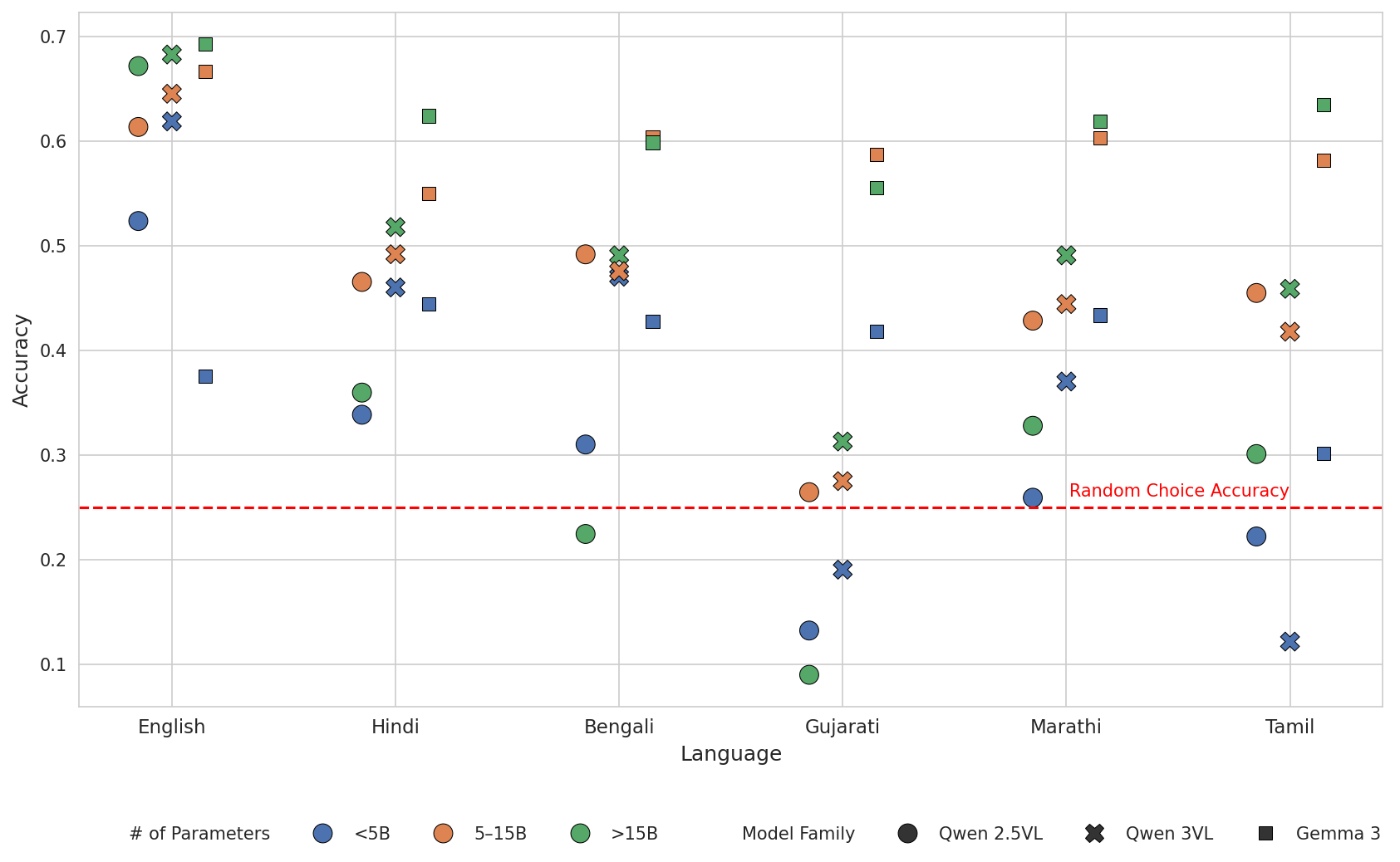}
    \caption{Zero-shot Inference}
\end{subfigure}

\vspace{0.25cm}

\begin{subfigure}[t]{0.48\textwidth}
    \centering
    \includegraphics[width=\linewidth]{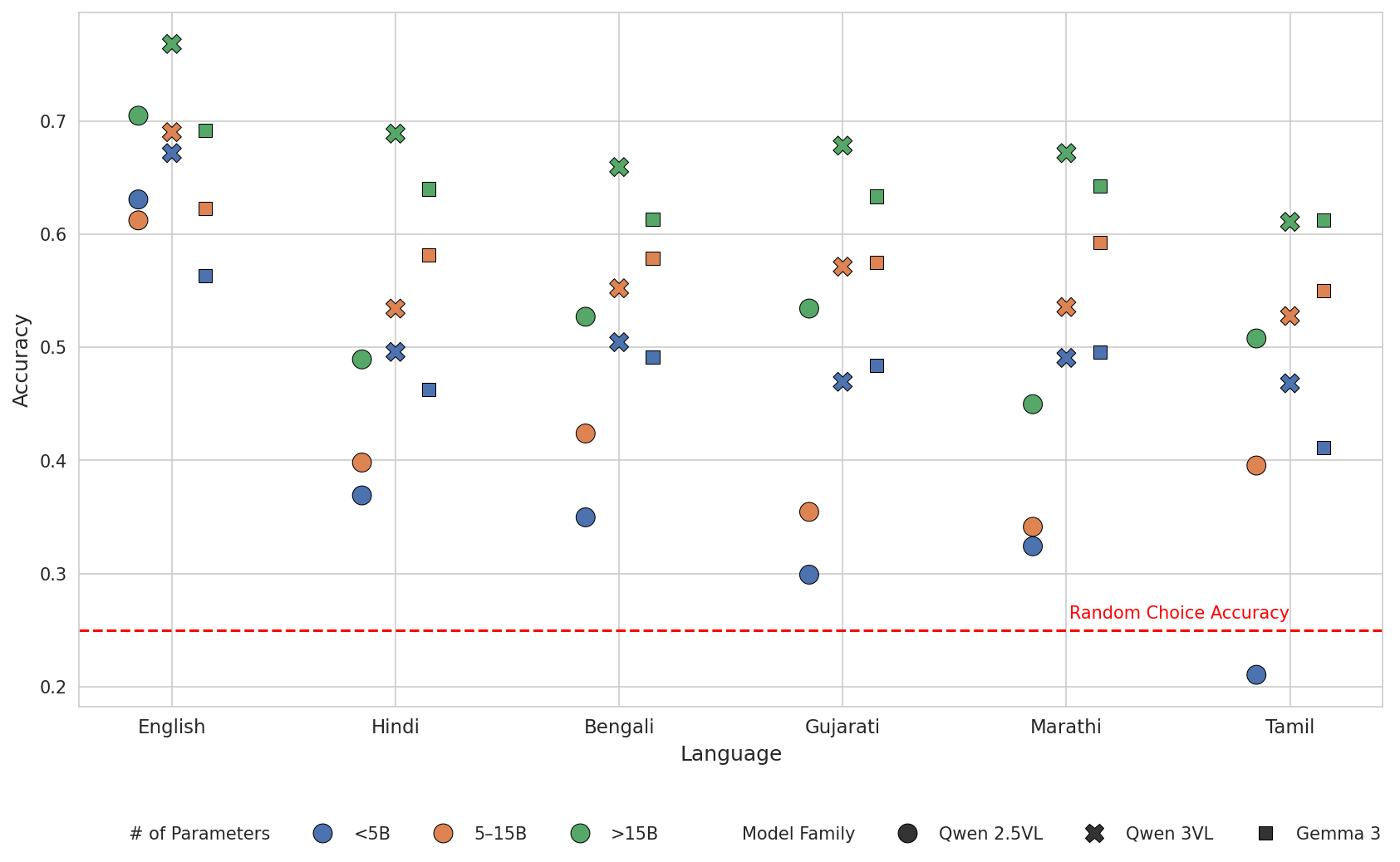}
    \caption{Constrained decoding}
\end{subfigure}
\hfill
\begin{subfigure}[t]{0.48\textwidth}
    \centering
    \includegraphics[width=\linewidth]{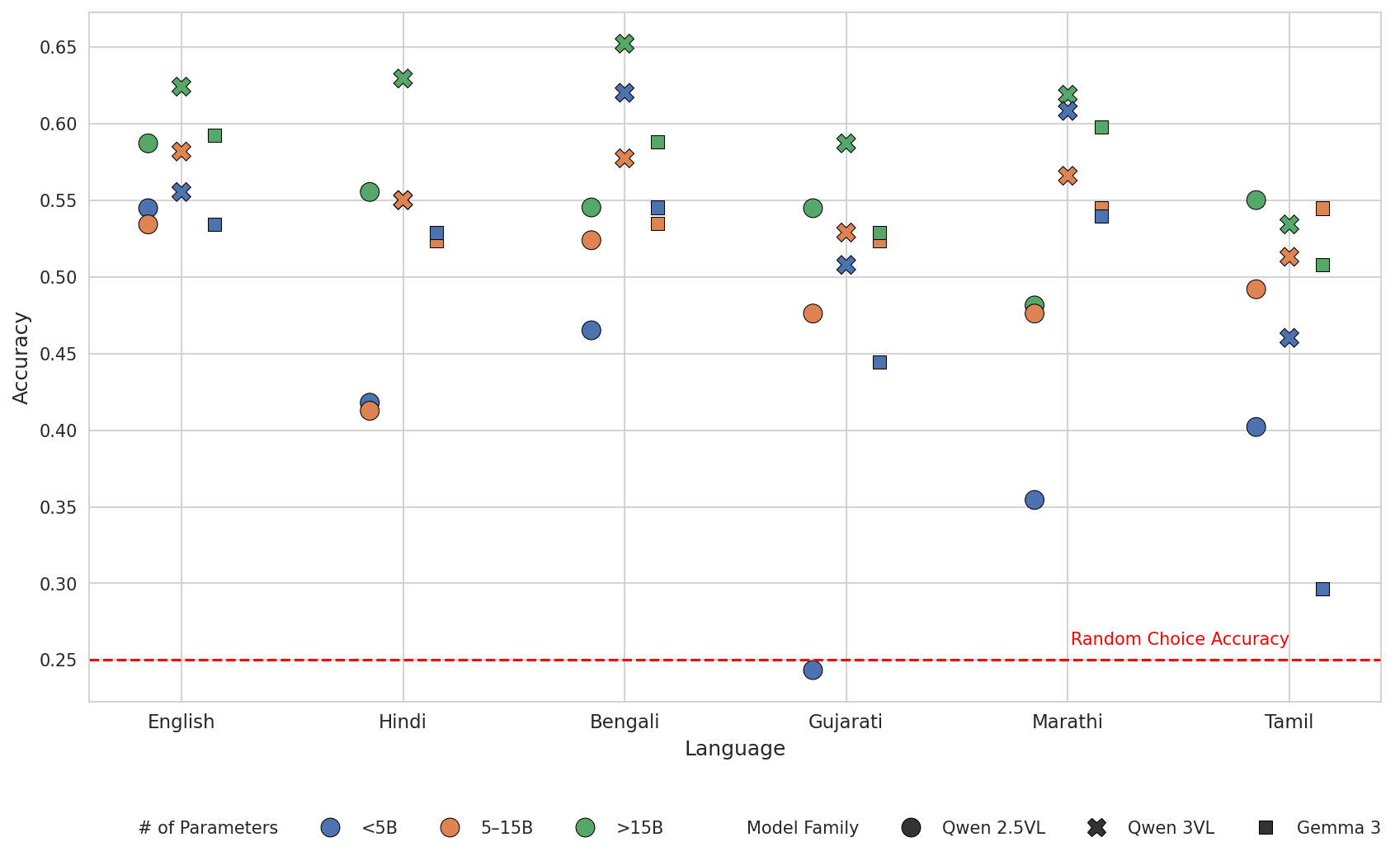}
    \caption{Constrained decoding}
\end{subfigure}

\vspace{0.25cm}



\begin{subfigure}[t]{0.48\textwidth}
    \centering
    \includegraphics[width=\linewidth]{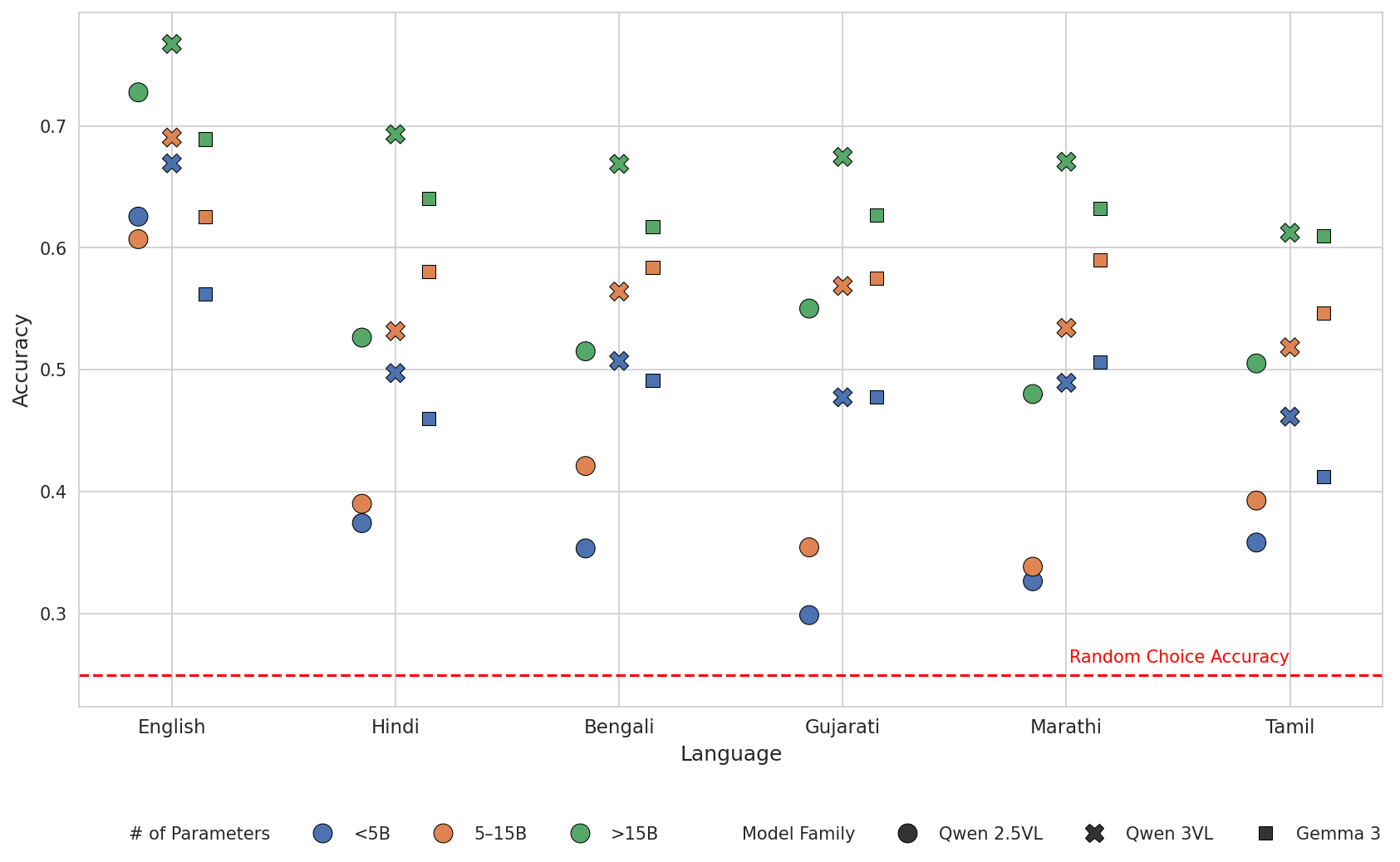}
    \caption{SFT + constrained decoding}
\end{subfigure}
\hfill
\begin{subfigure}[t]{0.48\textwidth}
    \centering
    \includegraphics[width=\linewidth]{latex/inference-and-sft-plots/option-2/SFT-img_ep03-multi_lingual-combined_metrics.png}
    \caption{SFT + constrained decoding}
\end{subfigure}

\caption{
Comparison of model performance across inference and training strategies.
Rows correspond to different generation strategy,
while columns contrast text-only and text+image inputs.
}
\label{fig:experiment_results_fig_option_2}
\end{figure*}

\begin{table*}
\centering
\scriptsize
\caption{Human evaluation scores across model families and parameter scales.}
\begin{tabular}{l|ccc|ccc|ccc}
\toprule
\textbf{Metrics} 
& \multicolumn{3}{c|}{\textbf{Qwen2.5VL}} 
& \multicolumn{3}{c|}{\textbf{Gemma 3}} 
& \multicolumn{3}{c}{\textbf{Qwen3VL}} \\
\cmidrule(lr){2-4} \cmidrule(lr){5-7} \cmidrule(lr){8-10}
& \textbf{3B} & \textbf{7B} & \textbf{32B}
& \textbf{4B} & \textbf{12B} & \textbf{27B}
& \textbf{4B} & \textbf{8B} & \textbf{32B} \\
\midrule
Financial Domain Understanding 
& 0.13 & 0.33 & 0.82 
& 0.24 & 0.43 & 0.91 
& 0.32 & 0.53 & 0.92 \\

Problem Interpretation \& Assumption Validity 
& 0.34 & 0.63 & 0.83 
& 0.54 & 0.73 & 0.84 
& 0.64 & 0.74 & 0.93 \\

Mathematical Correctness 
& 0.52 & 0.74 & 1.00 
& 0.63 & 0.84 & 1.00 
& 0.72 & 0.94 & 1.00 \\

Formula Selection \& Application 
& 0.63 & 0.74 & 1.00 
& 0.73 & 0.83 & 1.00 
& 0.73 & 0.83 & 1.00 \\

Reasoning Consistency and Answer Selection 
& 0.44 & 0.83 & 1.00 
& 0.63 & 1.00 & 1.00 
& 0.73 & 1.00 & 1.00 \\

Formatting \& Output Compliance 
& 0.74 & 1.00 & 1.00 
& 0.83 & 1.00 & 1.00 
& 0.83 & 1.00 & 1.00 \\
\midrule
\textbf{Average Score} 
& 0.46 & 0.70 & 0.94 
& 0.60 & 0.80 & 0.96 
& 0.66 & 0.85 & 0.98 \\
\bottomrule
\end{tabular}
\label{tab:financial_reasoning_human_eval}
\end{table*}

\begin{table*}[t]
\caption{Responses generated by various models for an English sample from FinVQA}\label{tab:qualitative_sample}
\scriptsize
\setlength{\tabcolsep}{3pt}
\begin{tabular}{|p{\linewidth}|}
\hline
\begin{center} \includegraphics[width=0.95\columnwidth]{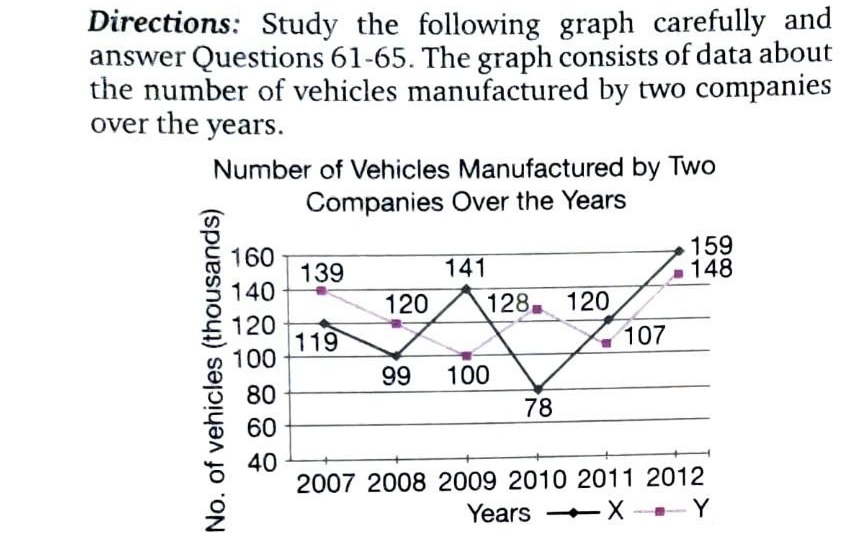} \\
\end{center}
\textbf{Question-} What is the difference between the total production of the two companies in the given years?\\
\textbf{Options-} A. 19,000 B. 22,000 C. 26,000 D. 28,000\\[2pt]
\textbf{Answer-} C\\[2pt]
\textbf{Reasoning:}\\
Total production of Company X from 2007 to 2012 in thousands = 119 + 99 + 141 + 78 + 120 + 159 \\= 716.\\
Total production of Company Y from 2007 to 2012 in thousands = 139 + 120 + 100 + 128 + 107 + 148 \\= 742.\\
Difference = 742 - 716 thousands = 26,000; so the correct option is C.
\\
\hline
\textbf{Response by Qwen 2.5 VL 3B -}\\
\textless{}reasoning\textgreater\\
First, I need to extract the data points for both companies from the graph and find their total sums.\\
Company X values: 119, 99, 141, 78, 120, 159.\\
Sum of X = 119 + 99 + 141 + 78 + 120 + 159 = 716.\\
Company Y values: 139, 120, 100, 128, 107, 148.\\
Sum of Y = 139 + 120 + 110 + 128 + 107 + 148 = 752.\\
Next, find the difference between the two totals: 752 - 716 = 36.\\
The y-axis is labeled "No. of vehicles (thousands)", so the actual difference is 34 * 1000 = 34,000. This is closest to the option D.\\
\textless{}/reasoning\textgreater\\
\textless{}answer\textgreater D \textless{}/answer\textgreater
\\
\hline

\textbf{Response by Qwen 3 VL 7B -}\\
\textless{}reasoning\textgreater\\
To find the total difference between the two companies, an alternative to summing all values first is to calculate the net difference\\ year by year (Company Y - Company X) and then sum those differences:\\
2007: 139 - 119 = +20\\
2008: 120 - 99 = +21\\
2009: 100 - 141 = -41\\
2010: 128 - 78 = +50\\
2011: 107 - 120 = -11\\
2012: 148 - 159 = -11\\
Now, sum these annual differences: 20 + 21 - 41 + 50 -11 - 11 = 26.\\
According to the vertical axis, the numbers are in thousands. Therefore, a difference of 26 corresponds to 26,000 vehicles.\\
\textless{}/reasoning\textgreater\\
\textless{}answer\textgreater Option C \textless{}/answer\textgreater
\\
\hline

\textbf{Response by Gemma 3 27B -}\\
\textless{}reasoning\textgreater\\
Based on the provided line graph, we must determine the total production for Company X and Company Y over the six-year period \\from 2007 to 2012, calculate the absolute difference between these totals, and adjust for the units. Note the y-axis explicitly states \\"No. of vehicles (thousands)".\\
Step 1: Calculate the total production for Company X (represented by the black line with diamond markers).\\
2007: 119\\
2008: 99\\
2009: 141\\
2010: 78\\
2011: 120\\
2012: 159\\
Total for Company X = 119 + 99 + 141 + 78 + 120 + 159 = 716 (in thousands).\\
Step 2: Calculate the total production for Company Y (represented by the purple line with square markers).\\
2007: 139\\
2008: 120\\
2009: 100\\
2010: 128\\
2011: 107\\
2012: 148\\
Total for Company Y = 139 + 120 + 100 + 128 + 107 + 148 = 742 (in thousands).\\
Step 3: Find the difference between the two total productions.\\
Difference = Total Y - Total X = 742 - 716 = 26 (in thousands).\\
Step 4: Convert to the final numerical value. 26 thousands is equal to 26,000. Checking the options provided, 26,000 corresponds to\\ Option C.\\
\textless{}/reasoning\textgreater\\
\textless{}answer\textgreater C \textless{}/answer\textgreater
\\
\hline
\end{tabular}
\end{table*}

\begin{table*}[t]
\centering
\caption{Multilingual Error Analysis for a  graph-based financial reasoning sample}
\label{tab:error_multilingual}
\scriptsize
\setlength{\tabcolsep}{3pt}
\begin{tabular}{|p{0.96\linewidth}|}
\hline
\begin{center}
\includegraphics[width=0.6\linewidth]{latex/Analysis.jpeg}\\
\end{center}
\textbf{Question-} What is the difference between the total production of the two companies in the given years?\\
\textbf{Options-} A. 19,000 \quad B. 22,000 \quad C. 26,000 \quad D. 28,000\\
\textbf{Answer-} C\\
\textbf{Reasoning-} Total production of Company X from 2007 to 2012 in thousands = 119 + 99 + 141 + 78 + 120 + 159 = 716. Total production of Company Y from 2007 to 2012 in thousands = 139 + 120 + 100 + 128 + 107 + 148 = 742. Difference = 742 - 716 thousands = 26,000; so the correct option is C.\\
\hline
\textbf{Bengali} \\[2pt]
\textless{}reasoning\textgreater \\
{\bengalifont

আমাদের কোম্পানি $X$ এবং কোম্পানি $Y$-এর মোট উৎপাদনের পার্থক্য হিসাব করতে হবে। প্রথম ধাপে, কোম্পানি $X$ এর মোট উৎপাদন যোগ করি: 119 + 99 + 141 + 78 + 120 + 159 = 726। দ্বিতীয় ধাপে, কোম্পানি $Y$ এর মোট উৎপাদন যোগ করি: 139 + 120 + 100 + 128 + 107 + 148 = 754। এখন, এই দুটির মধ্যে পার্থক্য বের করি: 754 - 726 = 28। $Y$-অক্ষ নির্দেশ করে যে মানগুলি হাজারে রয়েছে। তাই মোট পার্থক্য 28,000।}\\
\textless{}/reasoning\textgreater\\
\textless{}answer\textgreater D \textless{}/answer\textgreater
\\
\hline

\textbf{Hindi} \\[2pt]
\textless{}reasoning\textgreater \\
{\hindifont
ग्राफ़ के अनुसार, हमें कंपनी $X$ और कंपनी $Y$ का कुल उत्पादन निकालना होगा। कंपनी $X$ का कुल उत्पादन = 119 + 99 + 141 + 78 + 120 + 159 = 716. कंपनी $Y$ का कुल उत्पादन = 139 + 120 + 100 + 128 + 107 + 148 = 742. अब दोनों का अंतर निकालते हैं: 742 - 716 = 16. चूँकि संख्याएँ हजारों में हैं, इसलिए यह 16,000 होगा। यह सीधे तौर पर किसी विकल्प से मेल नहीं खाता, लेकिन सबसे करीब विकल्प $ A $(19,000) है।}\\
\textless{}/reasoning\textgreater\\
\textless{}answer\textgreater A \textless{}/answer\textgreater
\\
\hline

\textbf{Marathi} \\[2pt]
\textless{}reasoning\textgreater \\
{\marathifont
दोन कंपन्यांच्या उत्पादनातील एकूण फरक शोधण्यासाठी मला डेटा जोडावा लागेल. ग्राफ वाचताना, मी मूल्ये जवळच्या संख्येपर्यंत पूर्ण करतो: कंपनी $X$ ची अंदाजे बेरीज = 120 + 100 + 140 + 80 + 120 + 160 = 720. कंपनी $Y$ ची अंदाजे बेरीज = 140 + 120 + 100 + 130 + 110 + 150 = 750. दोन्हीमधील फरक = 750 - 720 = 30. ग्राफमधील मूल्ये हजारात असल्याने, हा फरक 30,000 असायला हवा. दिलेल्या पर्यायांमध्ये 28,000 हा सर्वात जवळचा पर्याय आहे.}\\
\textless{}/reasoning\textgreater\\
\textless{}answer\textgreater D \textless{}/answer\textgreater
\\
\hline

\textbf{Gujarati} \\[2pt]
\textless{}reasoning\textgreater \\
{\gujaratifont
કંપની $X$ અને કંપની $Y$ ના ઉત્પાદનનો તફાવત શોધવાનો છે. હું માત્ર શરૂઆતના અને છેલ્લા વર્ષનો તફાવત જોઈશ. 2007 માં તફાવત: 139 - 119 = 20. 2012 માં તફાવત: 159 - 148 = 11. જો આપણે આ સરેરાશ તફાવતો ઉમેરીએ, તો તે આશરે 19 થાય છે. વાય-અક્ષ મુજબ આ મૂલ્યો હજારોમાં છે, તેથી જવાબ 19,000 છે. આ વિકલ્પ $A$ સાથે મેળ ખાય છે.}\\
\textless{}/reasoning\textgreater\\
\textless{}answer\textgreater A \textless{}/answer\textgreater
\\
\hline

\textbf{Tamil} \\[2pt]
\textless{}reasoning\textgreater \\
{\tamilfont
இரண்டு நிறுவனங்களின் மொத்த உற்பத்தியின் வித்தியாசத்தை நாம் கணக்கிட வேண்டும். நிறுவனம் $X$-இன் தரவுகள்: 119, 99, 141, 78, 120, 159. இதன் மொத்த மதிப்பு = 716. நிறுவனம் $Y$-இன் தரவுகள்: 139, 120, 100, 128, 107, 148. இதன் மொத்த மதிப்பு = 742. இப்போது வித்தியாசத்தை காண வேண்டும்: 742 - 716 = 22. வரைபடத்தில் மதிப்புகள் ஆயிரங்களில் ($thousands$) கொடுக்கப்பட்டுள்ளன. எனவே, சரியான வித்தியாசம் 22,000 ஆகும்.}\\
\textless{}/reasoning\textgreater\\
\textless{}answer\textgreater B \textless{}/answer\textgreater
\\
\hline

\end{tabular}
\end{table*}

\section{LLM usage}
We used large language models (LLMs) to assist with code development and minor editing of the final manuscript.
\end{document}